\documentclass[3p, table]{elsarticle}
\usepackage[utf8]{inputenc} 
\usepackage[T1]{fontenc} 
\usepackage[%
  xindy,
  acronym,
  nopostdot,
]{glossaries}
\usepackage{csvsimple}
\usepackage{tabularx}
\usepackage{graphicx}

\usepackage[colorinlistoftodos]{todonotes}
\presetkeys{todonotes}{color=olive!40, size=\scriptsize}{}

\usepackage{lineno,hyperref}
\usepackage[nameinlink, capitalise, ]{cleveref}
\usepackage{pdflscape}
\modulolinenumbers[5]

\usepackage{booktabs}
\usepackage{array}
\usepackage[gen]{eurosym}
\usepackage{multirow}
\usepackage{color,soul}
\usepackage{footnote}
\usepackage{longtable}
\usepackage{threeparttable}
\usepackage{lscape}
\usepackage[usestackEOL]{stackengine}
\usepackage{placeins}
\usepackage[inline, shortlabels]{enumitem}
\usepackage{amsmath,amsfonts,amsthm,bm} 
\usepackage{adjustbox}
\usepackage{dtklogos}
\usepackage{tikz}
\usetikzlibrary{shapes.geometric, arrows.meta, positioning, fit, backgrounds, calc, shadows}
\usepackage{diagbox}
\usepackage{makecell}
\usepackage{subcaption}

\hypersetup{
  bookmarks=true,         
  unicode=false,          
  pdftoolbar=true,        
  pdfmenubar=true,        
  pdffitwindow=false,     
  pdfstartview={FitH},    
  pdftitle={},    
  pdfauthor={Author},     
  pdfsubject={Subject},   
  pdfcreator={Creator},   
  pdfproducer={Producer}, 
  pdfkeywords={keyword1} {key2} {key3}, 
  pdfnewwindow=true,      
  colorlinks=true,       
  linkcolor=red,          
  citecolor=blue,        
  filecolor=magenta,      
urlcolor=blue}           

\newcolumntype{P}[1]{>{\centering\arraybackslash}p{#1}}
\newcolumntype{C}[1]{>{\centering\arraybackslash}m{#1}}
\newcolumntype{M}{X<{\vspace{4pt}}}

\journal{}

\biboptions{authoryear}
\patchcmd{\emailauthor}{(#2)}{}{}{}

\graphicspath{{./plots}{../plots}}

\usepackage{subfiles}
\newacronym{tap}{TAP}{Traffic Assignment Problem}
\newacronym{dta}{DTA}{Dynamic Traffic Assignment}
\newacronym{ann}{ANN}{Artificial Neural Network}
\newacronym{gnn}{GNN}{Graph Neural Network}
\newacronym{taz}{TAZ}{Traffic Analysis Zone}
\newacronym{od}{OD}{Origin-Destination}
\newacronym{ue}{UE}{User Equilibrium}
\newacronym{so}{SO}{System Optimum}
\newacronym{bpr}{BPR}{Bureau of Public Roads}
\newacronym{vdf}{VDF}{Volume Delay function}
\newacronym{vcr}{V/C}{Volume-to-Capacity Ratio}
\newacronym{dag}{DAG}{Directed Acyclic Graph}
\newacronym{oba}{OBA}{Origin-Based Algorithm}
\newacronym{mlp}{MLP}{Multi-Layer Perceptron}
\newacronym{cnn}{CNN}{Convolutional Neural Network}
\newacronym{mpnn}{MPNN}{Message Passing Graph Neural Network}
\newacronym{gcn}{GCN}{Graph Convolutional Network}
\newacronym{gat}{GAT}{Graph Attention Network}
\newacronym{pinn}{PINN}{Physics Informed Neural Network}
\newacronym{aon}{AON}{All-Or-Nothing}
\newacronym{inn}{INN}{Implicit Neural Network}
\newacronym{cignn}{CI-SGNN}{City-Independent Spatial Graph Neural Network}
\newacronym{osm}{OSM}{OpenStreetMap}
\newacronym{poi}{POI}{Point of Interest}
\newacronym{rbf}{RBF}{Radial Basis Function}
\newacronym{vencoder}{V-Encoder}{Virtual Encoder}
\newacronym{rencoder}{R-Encoder}{Real Encoder}
\newacronym{tnrc}{TNRC}{Transportation Network for Research Core Team}
\newacronym{bfw}{BFW}{Bi-Conjugate Frank-Wolfe}

\newacronym{rmse}{RMSE}{Root Mean Square Error}
\newacronym{rmsn}{RMSN}{Normalized Root Mean Square Error}
\newacronym{mae}{MAE}{Mean Absolute Error}
\newacronym{fcn}{FCN}{Normalized Flow Conservation Error}

\newacronym{an-a}{AN-A}{Anaheim Dataset A}
\newacronym{an-b}{AN-B}{Anaheim Dataset B}
\newacronym{ch-a}{CH-A}{Chicago Sketch Dataset A}
\newacronym{ch-b}{CH-B}{Chicago Sketch Dataset B}

\newacronym{hetgat}{HetGAT}{Heterogeneous Graph Attention Network}
\newacronym{layer}{GUIDED}{Geometrically Unconstrained Inductive Demand EmbeDding}
\newacronym{model}{GUIDED-HetGAT}{GUIDED-Enhanced Heterogeneous Graph Attention Network}

\let\today\relax
\makeatletter
\def\ps@pprintTitle{%
    \let\@oddhead\@empty
    \let\@evenhead\@empty
    \def\@oddfoot{\footnotesize\itshape
         {} \hfill\today}%
    \let\@evenfoot\@oddfoot
    }
\makeatother
\begin{document}
\begin{frontmatter}

  \title{GUIDED Network-Agnostic Feature Initialization for Spatial Transferability in GNN-based Models}

  \author[TUMAddress]{Alessandro Scalese}
  \author[TUMAddress]{\corref{cor1}Santhanakrishnan Narayanan}
  \author[TUMAddress]{Constantinos Antoniou}
  \address[TUMAddress]{Technical University of Munich, Arcisstrasse 21, 80333 Munich, Germany}

  \cortext[cor1]{Corresponding author}\ead{santhanakrishnan.narayanan@tum.de}

  \begin{abstract}
    The macroscopic \acrlong{tap} is a fundamental but computationally expensive component of transportation planning. While \acrlong{gnn}s have emerged as fast, data-driven surrogates, their practical deployment is severely constrained by a spatial generalization gap. Standard models rely on transductive feature initializations that tie travel demand to fixed network topologies, preventing seamless transfer to new urban environments. To overcome this structural limitation, this research proposes a network-agnostic initialization layer, termed \gls{layer}. By injecting travel demand as a scalar attribute on auxiliary virtual links rather than as specific node features, this modular framework standardizes the input space regardless of network scale or the number of active origin-destination pairs.
    Extensive experimental evaluation across multiple urban topologies demonstrates that a \gls{hetgat} model integrated with the proposed \gls{layer} layer maintains state-of-the-art predictive accuracy on single-network tasks, while demonstrating superior robustness to out-of-distribution demand patterns and maintaining a distinct performance advantage over the baseline even under severe data scarcity.
    Notably, the proposed feature initialization enables highly parameter-efficient domain adaptation for inter-network transfer learning without artificial input homogenization, establishing a robust foundation for truly inductive models. At the same time, the optimized scatter operations of the initialization layer yield an approximate 50\% reduction in training time per epoch compared to the baseline approach.
    Furthermore, while demonstrated on vehicular traffic, this fundamental abstraction of spatial topology provides a versatile blueprint for generalized origin-destination spatial interaction problems, such as freight logistics and multimodal network optimization.
  \end{abstract}

\end{frontmatter}








\section{Introduction} \label{ch:intro}
Traffic simulation models are invaluable tools in traffic planning and management. As mobility demands continue to grow and evolve, city planners and decision-makers rely heavily on travel demand modeling to foresee traffic bottlenecks, evaluate infrastructure investments, and shape transportation policies.

A critical component of the planning process is the \gls{tap}, which seeks to determine the distribution of traffic flows across a network given a set of \gls{od} demands. Traditional approaches to solving the \gls{tap} rely on rigorous mathematical formulations and optimization techniques to find equilibrium solutions \citep{wardropRoadPaperTheoretical1952a,correaWardropEquilibria2011b,beckmann1956studies,patrikssonTrafficAssignmentProblem2015,de2024modelling}. However, these methods often rely on expensive iterative algorithms that do not scale well with the size and complexity of real-world networks \citep{patrikssonTrafficAssignmentProblem2015}, making \gls{tap} a significant bottleneck in modern strategic planning processes, limiting the ability of planners to explore suitable options and make informed decisions \citep{boyce2004convergence,patrikssonTrafficAssignmentProblem2015}.

In recent years, surrogate models based on machine learning approaches have emerged as promising alternatives to traditional \gls{tap} algorithms \citep{rahmani2023graph}. These models aim to learn the underlying patterns of traffic assignment from data, providing fast approximations of traffic flows without computing the full solution of the \gls{tap} for each scenario. In particular, \glspl{gnn} have shown great potential in capturing the complex relationships between network structure, demand patterns, and the resulting traffic flows, thanks to their ability to operate directly on graph-structured data \citep[e.g.,][]{rahmani2023graph,rahman2023data,narayanan2024graph,lassen2025learning}. Recent advancements have further enhanced their accuracy and reliability by embedding domain knowledge, in the form of physics-informed constraints, directly into the training process \citep{liu2024end}.

While these advancements demonstrate that \glspl{gnn} can provide fast and accurate traffic flow approximations on fixed network topologies, architectural constraints limit their practical utility in strategic transport planning. Practical deployment in planning workflows fundamentally relies on scenario analysis, where decision-makers must evaluate the macroscopic impacts of physical network modifications, such as capacity expansions, infrastructure disruptions, or structural road closures. However, current \gls{gnn}-based surrogates fail in these dynamic context variations due to structural limitations in their architecture or instantiation methods.
Most models in the literature map the travel demand matrix, whose dimensions are tied to a specific network, directly onto node feature layouts, resulting in intrinsically transductive models. Consequently, evaluating any network modification or transferring the surrogate to an alternative urban topology requires modifying the underlying model architecture, severely restricting its practical applicability and potential for transfer learning.

To address this limitation, this research introduces the \acrfull{layer}, a novel, modular, network-agnostic feature initialization framework designed for spatial transferability. Instead of relying on rigid matrix layouts, this approach re-frames demand initialization as a localized graph aggregation problem. By leveraging so-called virtual links in a heterogeneous graph, this procedure translates global travel demand (in the form of \gls{od} matrices) directly into localized, meaningful node and edge features. The systematic decoupling of the input feature dimensions from the physical scale of the specific network ensures that the dimensions of the input space remain strictly uniform across different networks, regardless of their size and topology. This standardization allows for straightforward transfer of surrogate models across different networks, removing the need for any structural modifications to the model. Crucially, this approach is also not bound to this specific application: any system reliant on the spatial interaction of origin-destination pairs can benefit from this network-agnostic initialization mechanism, leveraging it to achieve seamless spatial transferability.

To validate this approach, the proposed initialization procedure is integrated with a state-of-the-art heterogeneous \gls{gnn} architecture for traffic assignment, and its performance is evaluated on multiple real-world networks. The results demonstrate that the proposed framework effectively enables surrogate models to be transferable across diverse network topologies. On single-network tasks, the integration of the \gls{layer} initialization layer allows the model to maintain high predictive accuracy and surpass the baseline's robustness to out-of-distribution demand patterns. Even under severe data scarcity, the proposed layer grants the resulting model a distinct performance advantage. Notably, the framework also enables efficient domain adaptation, allowing cross-network transfer without artificial input homogenization.

The primary objectives of this work can therefore be summarized as:
\begin{itemize}
  \item Formulate a robust, network-agnostic initialization procedure for surrogate \gls{gnn} models.
  \item Integrate this procedure into an existing state-of-the-art surrogate \gls{gnn} architecture for traffic assignment.
  \item Evaluate the performance of the proposed approach on real-world networks to validate its predictive accuracy, physical consistency and spatial transferability.
\end{itemize}

To achieve this final objective, the experimental framework is structured into four distinct evaluation phases. First, intra-network generalization establishes the model's predictive accuracy and performance ceiling under standard conditions. Second, sample efficiency evaluations test the architecture's robustness to severe data scarcity during training. Third, semi-supervised inference assesses the model's capacity to reconstruct unobserved macroscopic network states under partial link observability. Finally, inter-network transfer learning validates the model's spatial generalizability across disparate urban topologies, utilizing layer-wise ablation to isolate transferable flow physics.

\begin{figure}[htb]
  \centering
  \resizebox{\textwidth}{!}{%
    \begin{tikzpicture}[
        node distance = 1.8cm and 2cm,
        mainbox/.style = {draw, rectangle, minimum width=4.5cm, minimum height=1.2cm, text width=4cm, align=center, fill=gray!15, font=\sffamily\small\bfseries, drop shadow={opacity=0.15}, execute at begin node={\hyphenpenalty=10000 \exhyphenpenalty=10000 \relax}},
        subbox/.style = {draw, rectangle, minimum width=4.5cm, minimum height=1.2cm, text width=4cm, align=center, fill=white, font=\sffamily\small, drop shadow={opacity=0.1}, execute at begin node={\hyphenpenalty=10000 \exhyphenpenalty=10000 \relax}},
        smallbox/.style = {draw, rectangle, minimum width=3.4cm, minimum height=1.1cm, text width=3cm, align=center, fill=white, font=\sffamily\footnotesize, drop shadow={opacity=0.1}, execute at begin node={\hyphenpenalty=10000 \exhyphenpenalty=10000 \relax}},
        doublearrow/.style = {{Stealth[length=2.5mm]}-{Stealth[length=2.5mm]}, thick},
        arrow/.style = {-{Stealth[length=2.5mm]}, thick},
        dashedarrow/.style = {-{Stealth[length=2.5mm]}, thick, densely dashed, color=black!70},
        dashedline/.style = {thick, densely dashed, color=black!60}
      ]

      \node[mainbox] (obj) {Research Objectives};
      \node[mainbox, below=1.75cm of obj] (lit) {Literature Review};
      \node[mainbox, below=1.75cm of lit] (meth) {Methodology};
      \node[mainbox, below=1.75cm of meth] (exp) {Experiments \& Results};
      \node[mainbox, below=1.75cm of exp] (concl) {Conclusions};

      \draw[doublearrow] (obj) -- (lit);
      \draw[arrow] (lit) -- (meth);
      \draw[arrow] (meth) -- (exp);
      \draw[arrow] (exp) -- (concl);

      \node[subbox, left=2.2cm of lit, yshift=2cm] (limgnn) {Limitations of Current\\GNN-TAP Surrogates};
      \node[subbox, below=0.6cm of limgnn] (spatgap) {Spatial Generalizability Gap\\(Topology \& Demand Mismatch)};

      \draw[arrow] (limgnn) -- (spatgap);

      \begin{scope}[on background layer]
        \node[draw, densely dashed, inner sep=0.4cm, fill=gray!5, fit=(limgnn) (spatgap)] (lit_group) {};
      \end{scope}

      \draw[dashedline] (lit_group.north east) -- (lit.north west);
      \draw[dashedline] (lit_group.south east) -- (lit.south west);

      \node[subbox, left=2.2cm of meth, yshift=0.3cm] (hetgraph) {Heterogeneous Graph\\Construction};
      \node[subbox, below=0.6cm of hetgraph] (guided) {GUIDED Feature Initialization\\(Linear \& RBF Variants)};
      \node[subbox, below=0.6cm of guided] (hetgat) {Integration with\\HetGAT Architecture};

      \draw[arrow] (hetgraph) -- (guided);
      \draw[arrow] (guided) -- (hetgat);

      \begin{scope}[on background layer]
        \node[draw, densely dashed, inner sep=0.4cm, fill=gray!5, fit=(hetgraph) (guided) (hetgat)] (meth_group) {};
      \end{scope}

      \draw[dashedline] (meth_group.north east) -- (meth.north west);
      \draw[dashedline] (meth_group.south east) -- (meth.south west);

      \node[subbox, right=3.6cm of exp, yshift=7cm] (basenet) {
        \textbf{Base Networks}\\Anaheim \& Chicago
      };
      \node[subbox, below=0.4cm of basenet] (cappert) {
        \textbf{Capacity Perturbation}\\Unified Random Scaling
      };

      \node[smallbox, anchor=north east] (data) at ([xshift=-0.15cm, yshift=-0.6cm]cappert.south) {\textbf{Dataset A}\\Random Uniform Scaling};
      \node[smallbox, anchor=north west] (datb) at ([xshift=0.15cm, yshift=-0.6cm]cappert.south) {\textbf{Dataset B}\\Distribution Shift};

      \draw[arrow] (basenet.south) -- (cappert.north);
      \draw[arrow] (cappert.south) -- +(0, -0.2) -| (data.north);
      \draw[arrow] (cappert.south) -- +(0, -0.2) -| (datb.north);

      \begin{scope}[on background layer]
        \node[draw, dotted, thick, inner sep=0.2cm, fill=none, fit=(basenet) (cappert) (data) (datb)] (data_group) {};
      \end{scope}

      \coordinate (grid_top) at ([yshift=-1.8cm]data_group.south);
      \node[smallbox, anchor=south east] (expa) at ([xshift=-0.15cm, yshift=0cm]grid_top) {\textbf{Exp A}\\Intra-Network Generalization};
      \node[smallbox, anchor=south west] (expb) at ([xshift=0.15cm, yshift=0cm]grid_top) {\textbf{Exp B}\\Sample Efficiency};
      \node[smallbox, anchor=north east] (expc) at ([xshift=-0.15cm, yshift=-0.3cm]grid_top) {\textbf{Exp C}\\Semi-Supervised Inference};
      \node[smallbox, anchor=north west] (expd) at ([xshift=0.15cm, yshift=-0.3cm]grid_top) {\textbf{Exp D}\\Transfer Learning \& Ablation};

      \begin{scope}[on background layer]
        \node[draw, dotted, thick, inner sep=0.2cm, fill=none, fit=(expa) (expb) (expc) (expd)] (exp_tests) {};
      \end{scope}

      \coordinate (eval_grid_top) at ([yshift=-1.8cm]exp_tests.south);
      \node[smallbox, anchor=south east] (evala) at ([xshift=-0.15cm, yshift=0cm]eval_grid_top) {\textbf{Statistical Metrics}\\MAE, RMSN, $R^2$};
      \node[smallbox, anchor=south west] (evalb) at ([xshift=0.15cm, yshift=0cm]eval_grid_top) {\textbf{Domain Metrics}\\\% GEH < 5, FCN};

      \begin{scope}[on background layer]
        \node[draw, densely dashed, inner sep=0.4cm, fill=gray!5, fit=(basenet) (cappert) (data) (datb) (expa) (expb) (expc) (expd) (evala) (evalb)] (exp_group) {};

        \node[draw, dotted, thick, inner sep=0.2cm, fill=white, fit=(expa) (expb) (expc) (expd)] (exp_tests) {};
        \node[draw, dotted, thick, inner sep=0.2cm, fill=white, fit=(basenet) (cappert) (data) (datb)] (data_group) {};
        \node[draw, dotted, thick, inner sep=0.2cm, fill=white, fit=(evala) (evalb)] (eval_group) {};
      \end{scope}

      \draw[arrow] (data_group.south) -- (exp_tests.north);
      \draw[arrow] (exp_tests.south) -- (eval_group.north);

      \draw[dashedline] (exp.north east) -- (exp_group.north west);
      \draw[dashedline] (exp.south east) -- (exp_group.south west);

    \end{tikzpicture}
  }
  \caption{Overview of the research structure and methodological framework, mapping the top-down progression of the study.}
  \label{fig:methodology_flowchart}
\end{figure}
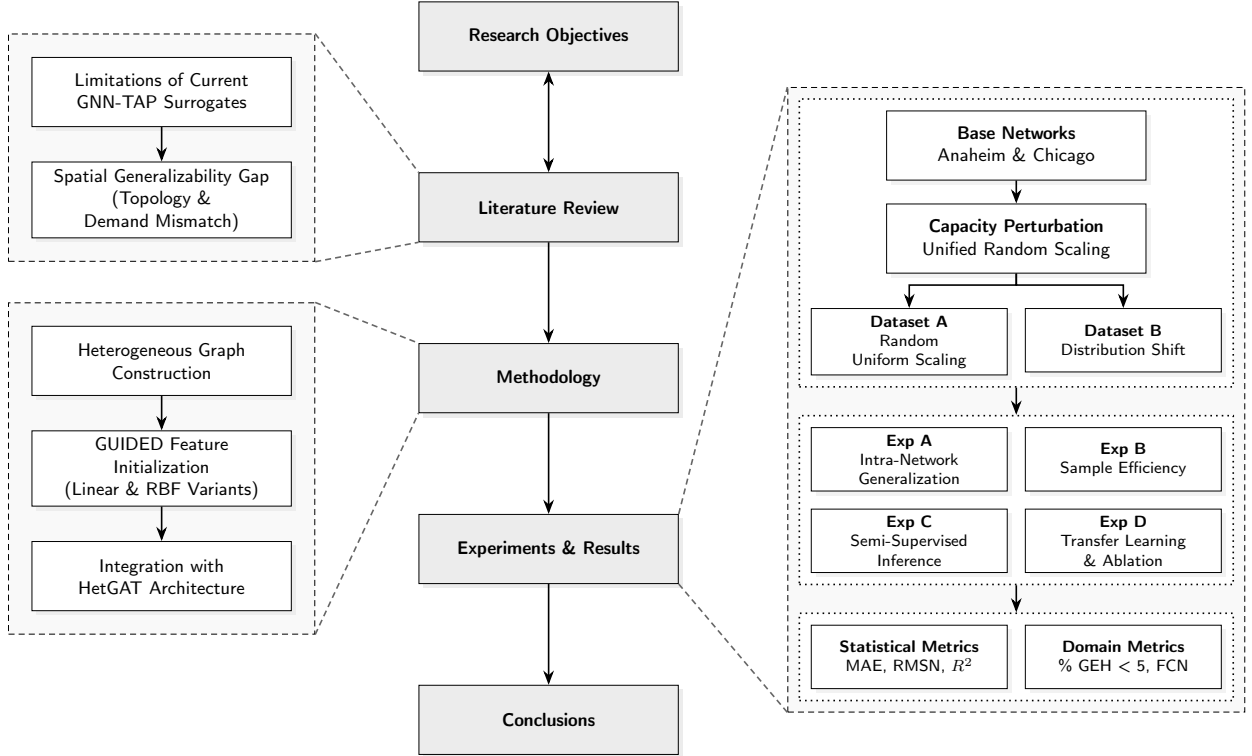

To illustrate the overarching structure and contributions of this research, \cref{fig:methodology_flowchart} maps the top-down progression of the study, from the identification of the spatial generalization gap to the final conclusions, while the lateral expansions detail the core contributions of each section. The remainder of this article is structured as follows. \cref{ch:literature} reviews state-of-the-art applications of \glspl{gnn} to the \gls{tap} and previous spatial generalization approaches. \cref{ch:methodology} details the proposed network-agnostic initialization procedure and its integration with the surrogate \gls{gnn} architecture. \cref{ch:experiment-setup} presents the experimental setup, while the results of the performance evaluations and their implications are discussed in \cref{ch:results}. Finally, \cref{ch:conclusion} summarizes the core findings of this research, acknowledges its methodological limitations, and outlines promising directions for future work.

The entire codebase, including data generation, processing scripts, and model implementations, is provided in a private GitHub repository, which is found at \url{https://github.com/scal-o/guided-framework}. Access to the repository is available upon request.

\section{Literature Review} \label{ch:literature}
The literature on the application of \glspl{gnn} to the \gls{tap} has seen a rapid evolution in recent years, transitioning from basic adaptations of standard architectures to highly specialized models designed to capture the unique characteristics of transportation networks and traffic flow dynamics. In the following sections, we review the existing literature on the use of \glspl{gnn} as surrogate models for traffic assignment (\cref{sec:gnn-for-tap}), analyze current methodological constraints regarding spatial generalization to unseen urban networks (\cref{sec:gnn-generalization}), and finally highlight the research gaps that motivate the proposed contribution of this work (\cref{sec:research-gaps}).

\subsection{\acrlongpl{gnn} for the \acrlong{tap}} \label{sec:gnn-for-tap}
As \cite{borde2025mathematical} note, the fundamental objective of a \gls{gnn} is to learn a signal over non-Euclidean graph structures, which involves capturing and representing the complex relationships and dependencies between nodes, edges, and the overall graph topology. Different \gls{gnn} formulations approach this problem from different perspectives: \glspl{gcn} generalize the concept of convolution to graph-structured data \citep{kipf2016semi}; spectral methods rely on the spectral decomposition of the graph Laplacian to define convolution operations in the frequency domain \citep{bruna2013spectral}; graph transformers extend the transformer architecture \citep{vaswani2017attention} to the graph domain, allowing for direct interactions between all nodes in a graph, regardless of their topological distance \citep{dwivedi2020generalization}; \glspl{mpnn} define convolution-like operations directly in the spatial domain of the graph, where nodes iteratively exchange and aggregate messages from their local neighborhoods \citep{gilmer2017neural,gilmer2020message,borde2025mathematical}.

In the context of the \gls{tap}, \cite{rahman2023data} first applied a \gls{gcn} architecture to the problem, formulating it as a supervised learning problem mapped onto a fixed graph structure. They successfully demonstrated that a \gls{gcn} could learn to approximate the equilibrium traffic flows on a given network with high accuracy, while significantly reducing the computational time compared to traditional algorithms.

Subsequent research from \cite{narayanan2024graph,makarov2024development} focused on developing proof-of-concept architectures acting as surrogates for the entire four-step model. These models, based on message-passing \gls{gat} \citep{velivckovic2017graph,brody2021attentive}, were designed to capture the complex interactions between the different stages of the four-step model and provide fast approximations of city-wide demand patterns. \cite{hu2025use} also incorporated the \gls{gat} architecture to develop an encoder-decoder framework for traffic assignment. Moreover, the path-based variant of their model enforces link-level flow conservation as a strict mathematical constraint via path-to-link incidence relationships, ensuring that the predicted traffic flows adhere to fundamental physical principles.

To model transport supply and demand interactions more accurately, \cite{liu2024end} proposed a \gls{mpnn} architecture based on Heterogeneous Graphs. They introduce the concept of auxiliary \textit{virtual links} connecting \gls{od} node pairs, capturing travel demand aspects in the topology of the network itself. This approach extends the concept of local neighborhood aggregation by enabling direct message passing between physically non-adjacent nodes, resulting in increased feature propagation efficiency.
Furthermore, rather than imposing hard topological constraints, they enforce node-based flow conservation as a soft penalty within the loss function, nudging the model towards physically realistic solutions.  Their subsequent work then extends the model using multi-graph views to model modal split choices alongside the traffic assignment mechanics \citep{liu2025multi}.

Moving beyond the Message Passing paradigm, \cite{lassen2025learning} incorporated global attention mechanisms into their \gls{gnn} architecture, leveraging a GraphGPS \citep{rampavsek2022recipe} structure based on Graph Transformers \citep{dwivedi2020generalization}. Their usage of global attention and network centroids, representing the nodes involved in \gls{od} pairs, leverages direct interactions between all nodes in the graph to capture long-range dependencies and complex interactions in the transportation network without being constrained by the local neighborhood structure or by the depth problems that arise in \gls{mpnn}. More recently, \cite{ameli2026optimization} proposed a sequence-to-sequence Transformer-based architecture to learn \gls{ue} directly in the path-flow space by leveraging global self-attention mechanisms. Both approaches resemble the virtual link paradigm introduced by \cite{liu2024end}, as they aim to model long-distance interactions between nodes in the transportation network, albeit through different architectural mechanisms.

Other recent advancements have focused on scalability and rapid adaptability. \cite{liu2025scalable} proposed a scalable learning approach utilizing \glspl{gcn} coupled with graph partitioning to handle massive, large-scale urban networks. Concurrently, \cite{agriesti2026metalearning} introduced meta-learning to traffic assignment, aiming to enable rapid adaptation of surrogate models to new, unseen traffic conditions, such as localized capacity drops or infrastructural disruptions.

\subsection{Spatial Generalization for \acrlongpl{gnn}} \label{sec:gnn-generalization}
Despite the significant advancements in predictive accuracy and computational efficiency, \gls{gnn}-based surrogate models remain constrained by a spatial generalization gap: they are unable to seamlessly transfer learned traffic dynamics to new and different urban topologies. Our research shows that this structural limitation stems from a two-fold input mismatch problem: a \textit{topological} mismatch and a \textit{demand feature} mismatch.

\subsubsection{The Topological Mismatch}
The topological mismatch arises from structural dependencies within the model architectures. Traditional configurations like the \gls{gcn} framework used by \cite{rahman2023data} rely on convolutional filters calibrated on specific graph sizes and topologies. Because this approach fundamentally relies on a static, graph-specific Laplacian matrix, the resulting model is strictly transductive and as such structurally incapable of processing different networks. \cite{liu2025scalable} attempted to circumvent this issue through algorithmic partitioning, allowing the model to handle variable topology networks by breaking them into fixed-size subgraphs and employing zero-padding to maintain matrix dimensional consistency. However, this method introduces significant trade-offs, as the model's limited exposure to only partitions of the graph during training prevents it from capturing global structural dependencies, leading to performance degradation when applied to larger networks requiring extensive partitioning (as noted by the authors themselves).

Recent efforts by \cite{agriesti2026metalearning} investigated a meta-learning framework based on a GatedGCN architecture \citep{bresson2017residual} to adapt to dynamic network conditions, such as road closures or localized capacity drops. While effective for localized topological adjustments, this approach operates under the assumption of a fixed global graph size, leaving the challenge of generalizing to unseen network scales and topologies unaddressed.

\subsubsection{The Demand Feature Mismatch}
While the topological mismatch can be managed using inductive paradigms such as \glspl{mpnn} and Graph Transformers, the demand feature mismatch is a challenge specific to transportation engineering applications and to the \gls{tap} in particular. In the \gls{tap}, the primary demand-side input is the localized travel demand, represented by an \gls{od} matrix whose size is determined by the number of \glspl{taz} and \gls{od} pairs in the network.

In most existing implementations, this matrix is mapped directly to node features, with each node containing a vector representing the travel demand originating from that node to all other nodes in the network \citep[e.g.][]{liu2024end,lassen2025learning,hu2025use}. Similarly, the Transformer-based approach by \cite{ameli2026optimization} flattens the transportation network features into a tensor where each row represents an \gls{od} pair. These formulations tie input feature dimensions directly to a specific graph structure: consequently, even inherently inductive paradigms fail on alternative networks as the new input feature dimensions do not match the ones established during the model training.

To allow their model to be applied to different networks, \cite{liu2025end} projected demand features into a fixed-size embedding space and relied on structural homogenization of the input space, padding the input matrix with dummy nodes up to a predefined maximum size. This approach allows for a consistent input dimension across different networks, but it also introduces limitations: it sets a rigid spatial ceiling, and introduces non-physical entities that can alter the model's hidden message-passing representations. Similarly, the Graph Transformer implementation proposed by \cite{lassen2025learning} also relies on a direct mapping of the \gls{od} matrix to node features.

To bypass raw demand matrix dependencies, \cite{liu2025scalable} use an \gls{aon} shortest-path algorithm to compute an initial traffic flow distribution on the network, encoded directly into the node embeddings of a dual graph representation. While this approach yields scale-agnostic input features, the \gls{aon} flow distribution does not capture the complex demand-supply interactions in the transportation network, and thus might not provide a sufficiently informative representation of traffic dynamics for the \gls{gnn} to learn true equilibrium states effectively.

\subsubsection{Generalization Approaches}
The pursuit of true inductive capabilities in traffic assignment remains an open challenge, with several distinct approaches emerging in the literature.

From a theoretical standpoint, \glspl{inn} represent an ideal inductive framework due to their architectural scale-invariance. Building on this property, \cite{liu2025end} introduced an end-to-end framework integrating \gls{inn} layers, specifically designed to enforce Wardrop's \gls{ue} conditions, to jointly estimate both demand and traffic equilibrium states from multiday traffic state observations. Their work also provides rigorous theoretical bounds on the model expressivity and generalization capabilities, demonstrating that, given sufficient data, the framework can transfer learned observation dynamics to unseen traffic conditions. However, their reliance on fixed-point iterations presents significant practical challenges, such as the need for complex custom auto-differentiation solvers.

For cross-city transfer, \cite{erdol2025city} developed the \gls{cignn} architecture. However, their design focuses on distinct classification objectives (predicting local demand for bike-sharing services), leveraging standardized, scale-agnostic topological features (such as the \gls{poi} data from \gls{osm}). This approach cannot be directly adapted to the context of \gls{tap} without a mechanism to integrate variable sized demand vectors as model features.

Alternatively, \cite{ju2025graphbridge} try to tackle the generalization gap from an architectural perspective through the introduction of the GraphBridge framework, pursuing arbitrary transfer learning across diverse graph topologies via side-tuning. Rather than retraining the underlying model for each new city, this framework introduces learnable auxiliary domain adapters to project input features and model outputs in the required graph- and task-specific spaces, respectively. This adapter-based approach, while architecturally sophisticated, attempts to reconcile dimensional mismatches in a \textit{post-hoc} fashion, adding computational complexity to "bridge" inherently transductive models rather than resolving the feature mismatch at its root.

\subsection{Summary of Research Gaps and Proposed Contribution} \label{sec:research-gaps}
The current state of the art reflects a spatial generalization gap in \gls{gnn}-based surrogate models for the \gls{tap}. While recent research has proposed promising solutions in the form of graph partitioning, post-hoc domain adaptation, and the integration of \gls{inn} layers, these approaches often result in a loss of global topological context or in additional implementation, deployment, and computational complexity of the resulting models.

Crucially, a distinction must be made between the structural capacity of a model's architecture and its input initialization procedures.
While modern paradigms such as \glspl{mpnn} and Graph Transformers are inherently inductive, their actual implementations fail on alternative networks because travel demand features are constructed transductively via rigid matrix layouts tied to a fixed spatial scale \citep[e.g.][]{liu2024end,lassen2025learning}.

To address these limitations without modifying the already existing downstream architectures, this work introduces a modular, network-agnostic initialization procedure that dynamically projects travel demand into a standardized latent space. Leveraging the virtual link paradigm introduced by \cite{liu2024end}, we propose a two-step approach to demand features initialization:
\begin{itemize}
  \item First, raw scalar travel demand values are expanded into high-dimensional edge embeddings directly on the auxiliary virtual links connecting \gls{od} pairs;
  \item Second, a specialized, permutation-invariant aggregation layer pools these virtual edge features to dynamically construct the initial physical node embeddings.
\end{itemize}

As this aggregation relies entirely on local graph connectivity rather than global matrix dimensions, it systematically decouples the initial input feature space from both the physical scale of the network and the semantic scale of the demand matrix. Furthermore, this procedure is designed to be architecturally agnostic. It functions as a standardized input-processing layer that can be implemented on top of existing inductive \gls{gnn} frameworks, enabling spatial generalization and domain adaptation across heterogeneous urban networks for already existing models.

\section{Methodology} \label{ch:methodology}
The literature review in \cref{ch:literature} highlighted the spatial generalization gap in current \gls{gnn}-based surrogate models for the \gls{tap}. This bottleneck is primarily driven by the structural inability of transductive feature initialization methods to handle varying network scales and demand feature dimensions. To overcome this limitation, this section details the mathematical formulation and architectural design of the proposed modular, network-agnostic initialization procedure, termed the \gls{layer} layer, and describes its integration within a state-of-the-art heterogeneous graph neural network architecture for traffic assignment.

\subsection{Construction of the Heterogeneous Graph} \label{sec:heterogeneous-graph}
The proposed framework is built upon a heterogeneous graph representation of the traffic network $\mathcal{G} = (\mathcal{V}, \mathcal{E}_r, \mathcal{E}_v)$, akin to the one introduced by \cite{liu2024end}, where:
\begin{description}
  \item[$\mathcal{V}$] is the set of vertices representing all physical intersections and \glspl{taz} within the network.
  \item[$\mathcal{E}_r$] is the set of directed real edges $r$, defining the actual road segments connecting the vertices in $\mathcal{V}$.
  \item[$\mathcal{E}_v$] is the set of directed virtual edges $q$, defining the auxiliary connections between \gls{od} pairs.
\end{description}

The real edges $r \in \mathcal{E}_r$ capture the physical topology of the network, and are associated with a static feature vector $\mathbf{e}_r \in \mathbb{R}^{2}$, containing  the normalized capacity and free-flow travel time. The virtual edges, on the other hand, represent an augmentation of the physical graph with demand information: each \gls{od} pair $(o,d)$ with a non-zero demand is connected by a directed virtual edge $q_{o,d}$, which serves as a structural routing conduit for the travel demand between the origin node $o$ and the destination node $d$.

\subsection{\gls{layer} Feature Initialization} \label{sec:guided-initialization}
The \gls{layer} initialization procedure consists of two sequential stages: first, the scalar travel demand associated with each virtual edge is projected into a high-dimensional embedding space, and second, these virtual edge embeddings are aggregated to construct the initial node embeddings. This two-stage process ensures that the initial node representations capture the global origin-destination interactions in a manner that is invariant to the specific network topology.

\subsubsection{Initialization of the Virtual Link Embeddings} \label{sec:initialization-virtual-embeddings}
The first stage of the proposed initialization procedure focuses on the virtual link embeddings. Each virtual edge $q_{o,d} \in \mathcal{E}_v$, as defined in \cref{sec:heterogeneous-graph}, is initially assigned a scalar attribute $d_q \in \mathbb{R}_+$ representing the travel demand between the origin node $o$ and the destination node $d$. The goal of this stage is to transform this scalar demand attribute into a high-dimensional embedding vector $\mathbf{e}_q \in \mathbb{R}^F$, where $F$ is the desired embedding dimension.

To evaluate the sensitivity of the model to the initial demand representation, we formulated and tested two distinct projection functions: a linear expansion featuring a standard \gls{mlp} architecture, and a \gls{rbf} expansion followed by a learnable linear projection.

\paragraph{Projection via Linear Expansion}
The first approach relies on a standard, configurable \gls{mlp} to project the scalar demand into the embedding space. The linear expansion for the case of a \gls{mlp} with a single layer can be formulated as:
\begin{equation} \label{eq:mlp-expansion}
  \mathbf{e}_q = MLP(d_q) = \sigma(\mathbf{W} d_q + \mathbf{b})
\end{equation}
where $\mathbf{W} \in \mathbb{R}^{F \times 1}$ is the learnable weight matrix, $\mathbf{b} \in \mathbb{R}^F$ is the learnable bias vector, and $\sigma$ is a non-linear activation function.

This formulation allows for a flexible mapping from the scalar demand to the embedding space, enabling the model to learn complex relationships between demand values and their corresponding embeddings.

\paragraph{Projection via \gls{rbf} Expansion}
The second approach utilizes a Gaussian \gls{rbf} expansion, followed by a linear layer to obtain the final embedding. \gls{rbf} expansion is a technique that decomposes a continuous scalar value into localized feature representations based on its (generalized) distance from a set of predefined centers. Given a set of $K$ centers $\{\mu_k\}_{k=1}^K$ and a bandwidth parameter $\sigma$, the Gaussian \gls{rbf} distance of the scalar demand $d_q$ from center $\mu_k$ can be expressed as:
\begin{equation} \label{eq:rbf-expansion}
  r_{q,k} = \exp\left(- \frac{(d_q - \mu_k)^2}{\sigma^2}\right)
\end{equation}
The resulting \gls{rbf} features $\mathbf{r}_q = [r_{q,1}, r_{q,2}, \dots, r_{q,K}]^\top$ are then linearly transformed to match the target embedding dimension:
\begin{equation} \label{eq:rbf-expansion-projection}
  \mathbf{e}_q = MLP(\mathbf{r}_q) = \sigma(\mathbf{W} \mathbf{r}_q + \mathbf{b})
\end{equation}
In this approach, each \gls{rbf} feature captures the relationship between the demand value and a specific region of the demand space, mapping complex demand distributions into linearly separable features for the downstream layers. The choice of centers and bandwidth can be tuned to capture relevant patterns in the demand data effectively.

Similar distance-based expansions have proven effective in capturing complex continuous spatial interactions in various domains \citep[e.g.][]{schutt2017schnet}, and such functional mappings have been shown to enhance the ability of neural networks to learn highly non-linear relationships from simple scalar inputs \citep[e.g.][]{tancik2020fourier}.

\subsubsection{Initialization of the Node Embeddings} \label{sec:initialization-node-embeddings}
The second stage of the proposed initialization procedure focuses on the node embeddings. To provide a standardized, network-agnostic input space, this stage leverages a permutation-invariant aggregation mechanism to pool information from virtual links directly into physical nodes. This abstracts global origin-destination interactions into localized, meaningful node embeddings $\mathbf{x}_v^{(0)}$ for each node $v \in \mathcal{V}$, ensuring spatial generalizability across disparate network topologies.

This initialization primarily involves the nodes that represent the centroids, or \glspl{taz}, of the network. Pure transit nodes, which are not the origin or destination of any trips, are simply initialized with a zero vector of the same dimension as the other node embeddings to maintain uniformity in the input space.

Centroid nodes can function as physical sources (trip generators) or sinks (trip attractors) for the traffic flow, or both. To explicitly preserve the directional nature of travel demand within the initial latent space, we perform two distinct, permutation invariant summation operations on the sets of incoming and outgoing virtual edges, respectively:
\begin{equation} \label{eq:outgoing-agg}
  \mathbf{x}_{out,v} = \sum_{q\;\in\;\mathcal{E}_v^{out,v}} \mathbf{e}_q
\end{equation}
\begin{equation} \label{eq:incoming-agg}
  \mathbf{x}_{in,v} = \sum_{q\;\in\;\mathcal{E}_v^{in,v}} \mathbf{e}_q
\end{equation}
Here, $\mathcal{E}_v^{out,v}$ and $\mathcal{E}_v^{in,v}$ denote the sets of outgoing and incoming virtual edges incident to node $v$, respectively. The resulting aggregated embeddings $\mathbf{x}_{out,v}$ and $\mathbf{x}_{in,v}$ are then concatenated to form the final node embedding:
\begin{equation} \label{eq:node-embedding}
  \mathbf{x}_v^{(0)} = [\mathbf{x}_{out,v} \;||\; \mathbf{x}_{in,v}]
\end{equation}
This approach ensures that the node embeddings capture the actual demand patterns associated with each node while maintaining a consistent embedding dimension across different networks.

\subsection{Integration into an Existing Model Architecture} \label{sec:model-integration}
The \gls{layer} layer described in the previous sections is intentionally designed as a modular preprocessing layer and is therefore independent of the downstream \gls{gnn} architecture.
To demonstrate this architectural flexibility, the proposed initialization procedure is integrated into the state-of-the-art \gls{hetgat} architecture introduced by \cite{liu2024end}.

The original \gls{hetgat} architecture consists of three sequential components:
\begin{enumerate}
  \item \textbf{Node Preprocessing}, which projects the high-dimensional node representations, derived directly from the \gls{od} matrix, into a fixed latent embedding space;
  \item \textbf{Heterogeneous Message Passing}, comprising a variable number of \gls{vencoder} layers followed by \gls{rencoder} layers. These attention-based layers propagate information across virtual \gls{od} connections and physical road links while maintaining separate message-passing mechanisms for the two edge types;
  \item \textbf{Edge Prediction}, where the learned node embeddings are combined with the physical edge attributes to predict the corresponding link \gls{vcr} through a dedicated \gls{mlp} prediction module.
\end{enumerate}

For a comprehensive reading on the model's architecture and mathematical formulations, the reader is referred to the original paper by \cite{liu2024end}.

The original implementation of the \gls{hetgat} model appends the \gls{od} matrix directly to the node features, assigning each node $i \in \mathcal{V}$ an input feature vector $\mathbf{x}_i \in \mathbb{R}^{|\mathcal{V}|}$, where $|\mathcal{V}|$ is the total number of nodes in the network, which are then projected into a fixed latent space in the node preprocessing stage.
The proposed \gls{layer} framework modifies only the initialization stage of the architecture. Rather than directly mapping the \gls{od} matrix to node features, travel demand is first embedded on the virtual edges and subsequently aggregated to construct the initial node embeddings, as described in \cref{sec:guided-initialization}. Once these embeddings have been initialized, the remaining computational pipeline, including the heterogeneous message-passing layers and the edge predictor, remains identical to the original \gls{hetgat} architecture.

This design has two important advantages. First, it demonstrates that the proposed initialization mechanism is modular and architecture-independent, allowing it to be incorporated into existing \gls{gnn} models without redesigning their internal operations. Second, by leaving the downstream architecture unchanged, it enables a controlled comparison with the baseline model in which any observed performance differences can be attributed exclusively to the proposed initialization strategy.

In the rest of this paper, we refer to the modified architecture as \gls{model}.
A high level comparative summary of the architectural differences and key hyperparameters for both the baseline \gls{hetgat} and the proposed \gls{model} implementations is provided in \cref{tab:architecture-summary}.

\begin{table}[!bp]
  \footnotesize
  \renewcommand{\arraystretch}{1}
  \begin{tabularx}{\textwidth}{@{} l M M @{}}
    \toprule
    \textbf{Component} & \textbf{Baseline (HetGAT)} & \textbf{Proposed (\acrshort{model})} \\
    \midrule
    Demand representation & OD matrix on nodes & OD scalar on virtual edges \\
    Node Embedding & Direct projection via MLP & Virtual edge aggregation via GUIDED Layer \hspace{2em} (\textit{lin} and \textit{rbf} variants)\\
    Latent dimension & 64 & Identical \\
    V-Encoder & 2 layers, 8 heads & Identical \\
    R-Encoder & 6 layers, 8 heads & Identical \\
    Edge Predictor & MLP ($130 \rightarrow 64 \rightarrow 1$) & Identical \\
    \bottomrule
  \end{tabularx}
  \captionof{table}{Summary of architecture components and key hyperparameters for the baseline and proposed models.}
  \label{tab:architecture-summary}
\end{table}

\subsubsection{\gls{model} Variants}
To evaluate the sensitivity to the initial demand representation, the \gls{layer} is instantiated using two distinct variants: \gls{model}-lin and \gls{model}-rbf. Both variants expand the scalar demand embedded on each virtual edge to a high-dimensional embedding $\mathbf{e}_q \in \mathbb{R}^{32}$ using either a standard \gls{mlp} or a \gls{rbf} expansion, as described in \cref{sec:guided-initialization}. For the linear variant of the model, this is achieved via a two-layer \gls{mlp} with a 16-unit hidden layer. For the \gls{rbf} variant, the scalar is projected onto 20 radial basis functions, with a shared learnable spread parameter $\sigma$, followed by a two-layer \gls{mlp} that maps the resulting 20-dimensional vector to the final 32-dimensional edge embeddings. For both variants, following the concatenation of incoming and outgoing virtual edges, the final aggregated node embeddings maintain a fixed dimensionality of $F=64$ to strictly align with the baseline HetGAT architecture.

\subsection{Physics-Informed Training Objective} \label{sec:loss-function}
To ensure the predicted traffic flows adhere to the fundamental physical principles of macroscopic traffic flow, the optimization objective is formulated as a composite $L_1$ penalty comprising three distinct terms, derived from \cite{liu2024end}. The total loss is defined as the weighted sum of the following components:
\begin{itemize}
  \item \textbf{V/C Loss} $\mathcal{L}_v$: primary supervised metric, evaluating the error between the predicted and ground-truth \gls{vcr} values.
  \item \textbf{Flow Loss} $\mathcal{L}_f$: secondary supervised metric, evaluating the error between the predicted flows (obtained by multiplying the predicted \gls{vcr} with the link capacities) and the ground-truth flows.
  \item \textbf{Conservation Loss} $\mathcal{L}_c$: physics-based constraint enforcing the principle of flow conservation at the node level, penalizing any violations across the network, encouraging the model to learn representations that adhere to the fundamental physical principles governing traffic flow.
\end{itemize}

For a given graph, let $\hat{y}_r$ represent the predicted \gls{vcr} value, $c_r$ the link capacity, and $\hat{f}_r = \hat{y}_r \cdot c_r$ the derived predicted flow for edge $r$. Let $y_r$ and $f_r$ denote the ground-truth \gls{vcr} and flow values for edge $r$, respectively. The total loss $\mathcal{L}$ is defined as:
\begin{equation} \label{eq:composite-loss}
  \mathcal{L} = \lambda_v \mathcal{L}_v + \lambda_{f} \mathcal{L}_f + \lambda_c \mathcal{L}_c
\end{equation}
where $\lambda_v$, $\lambda_f$, and $\lambda_c$ are the weights for the \gls{vcr} loss, flow loss, and conservation loss, respectively. The individual loss components are defined as follows:
\begin{align}
  \mathcal{L}_v &= \frac{1}{|\mathcal{E}_r|}\; \sum_{r\;\in\;\mathcal{E}_r}\;|\hat{y}_r - y_r| \\
  \mathcal{L}_f &= \frac{1}{|\mathcal{E}_r|}\; \sum_{r\;\in\;\mathcal{E}_r}\;|\hat{f}_r - f_r| \\
  \mathcal{L}_c &= \frac{1}{|\mathcal{V}|}\; \sum_{i\;\in\;\mathcal{V}}\;\left| \sum_{r\;\in\;\mathcal{E}_r^{out,i}} \hat{f}_r \; - \sum_{r\;\in\;\mathcal{E}_r^{in,i}} \hat{f}_r \; - \; d_{net,i} \right|
\end{align}
where $\mathcal{E}_r^{out,i}$ and $\mathcal{E}_r^{in,i}$ denote the sets of outgoing and incoming physical edges for node $i$, respectively, and $d_{net,i}$ is the net generated or attracted demand at node $i$.

\section{Experimental Setup} \label{ch:experiment-setup}
This section details the experimental framework designed to evaluate the efficacy, physical consistency, and spatial generalization capabilities of the proposed \gls{layer} initialization method for \glspl{gnn}. The experiments are structured to assess the models' performance across four distinct tasks:
\begin{enumerate}
  \item Intra-network generalization: evaluating the model's ability to approximate \gls{tap} solutions on the same network used for training.
  \item Intra-network generalization efficiency: evaluating the model's ability to approximate \gls{tap} solutions on the same network used for training when trained on a limited number of demand scenarios.
  \item Intra-network semi-supervised learning: evaluating the model's ability to approximate \gls{tap} solutions on the same network used for training when trained under partial observability of the network state.
  \item Inter-network transfer learning: evaluating the model's ability to approximate \gls{tap} solutions on a different network than the one used for training, with and without fine-tuning. The testing also includes an ablation study on the effectiveness of fine-tuning different layers of the model.
\end{enumerate}

This section provides a detailed description of the dataset generation process (\cref{sec:dataset-definition}) and of the training procedure (\cref{sec:training-procedure}), followed by a brief overview of the evaluation metrics (\cref{sec:evaluation-metrics}). Finally, the specific bounds and objectives of each experimental phase are detailed in \cref{sec:experiment-design}, where we discuss the rationale behind the experimental design choices and how they align with the overarching research objectives.

\subsection{Dataset Definition} \label{sec:dataset-definition}
This section describes the transportation networks and synthetic scenario generation procedures used to construct the datasets employed throughout the experimental evaluation. The datasets are designed to assess the models' ability to generalize across different traffic demand scenarios and network topologies, while also evaluating their physical consistency and predictive accuracy.

\subsubsection{Network Topologies}
\begin{figure}[hbtp]
  \centering
  \includegraphics[width=0.8\textwidth]{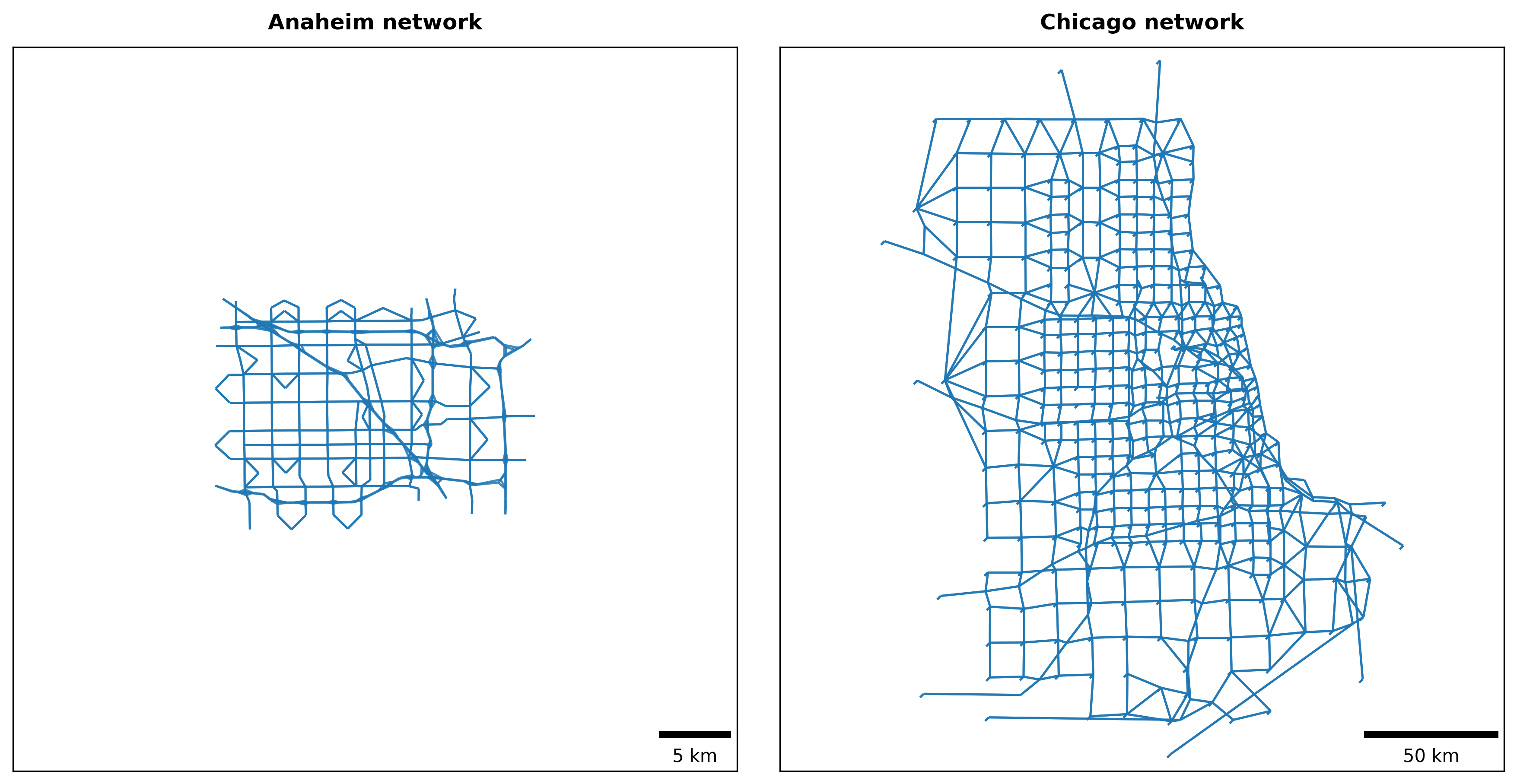}
  \captionof{figure}{Visual representation of the Anaheim (left) and Chicago Sketch (right) network topologies.}
  \label{fig:network-topologies}
\end{figure}

The experiments were conducted on two widely used urban transportation networks: the Anaheim network and the Chicago Sketch network. The original network topologies, demand matrices, link attributes, and other relevant data were obtained from the \gls{tnrc} repository \citep{Stabler2024TransportationNetworks}. A general overview of the networks' statistics is provided in \cref{tab:network-stats}, and a visual representation of the network topologies is shown in \cref{fig:network-topologies}.

The Anaheim network represents a medium-sized urban topology, located in Southern California. The network is characterized by a mix of arterial roads and highways, with varying link capacities and free-flow travel times. The Chicago Sketch network, on the other hand, represents a historical, aggregated, and larger urban topology, located in the city of Chicago. The network is characterized by a dense urban grid structure, with a high degree of connectivity and a wide range of link attributes, driven by the inclusion of both high-capacity freeway corridors and local, possibly residential streets.

\begin{table}[btp]
  \footnotesize
  \renewcommand{\arraystretch}{1}
  \begin{tabularx}{\textwidth}{@{} X r r r r r @{}}
    \toprule
    \textbf{Network} & \textbf{Nodes} & \textbf{Edges} & \textbf{Active OD Pairs} & \textbf{Avg. Degree} & \textbf{Total Demand} \\
    \midrule
    \begin{filecontents*}{network_summary_stats.csv}
network,numberofnodes,numberofedges,nonzeroods,averagenodedegree,totaldemand
Anaheim,416,914,1406,4.39,10469
Chicago,933,2950,46762,6.32,125335
\end{filecontents*}
\csvreader[
      head to column names,
      late after line=\\,
      late after last line=\\\bottomrule
    ]{network_summary_stats.csv}{}{%
      \network & \numberofnodes & \numberofedges & \nonzeroods & \averagenodedegree & \totaldemand%
    }%
  \end{tabularx}
  \captionof{table}{Summary statistics of the baseline transportation networks.}
  \label{tab:network-stats}
\end{table}

\subsubsection{Scenario Generation}
To adequately train and evaluate the \gls{gnn} ability to map topological and demand features to equilibrium flows, the generation of adequately diverse and representative datasets is crucial. For each of the networks, we generated two distinct datasets, each based on a different demand generation process and sharing a common capacity perturbation process.

For each dataset, we define an individual scenario as a unique combination of a demand matrix $\mathbf{D} \in \mathbb{R}^{|\mathcal{Z}| \times |\mathcal{Z}|}$, where $\mathcal{Z} \subset \mathcal{V}$ represents the set of \glspl{taz} defined in the network, and a capacity vector $\mathbf{c} \in \mathbb{R}^{|\mathcal{E}_r|}$. Together, the demand matrix and capacity vector uniquely define the input features and the corresponding \gls{tap} solution for a single scenario. By varying both the demand and capacity across scenarios, we can create a rich and diverse dataset that allows us to thoroughly evaluate the performance of our proposed \gls{gnn} model under a wide range of conditions.

Each dataset consists of $N=700$ scenarios, generated by applying a unified capacity perturbation process to the baseline network, while varying the demand generation process to create distinct datasets with different characteristics and levels of difficulty for the \gls{gnn} model. This dual-dataset structure is designed to test model performance under nominal, within-distribution daily variations (Dataset A), and under out-of-distribution, structurally decoupled demand profiles (Dataset B).

\paragraph{Capacity Perturbation}
The capacity perturbation process was designed to simulate localized infrastructure degradation, which is a common occurrence in real-world transportation networks due to factors such as construction, accidents, or maintenance activities. For each scenario $z \in \{1, \ldots, N\}$, the baseline capacity $c_r$ associated with link $r \in \mathcal{E}_r$, is perturbed following:
\begin{equation} \label{eq:capacity-generation}
  c_r^{(z)} = c_r \delta_r^{(z)}
\end{equation}
where $c_r^{(z)}$ is the perturbed capacity of link $r$ in scenario $z$ and $\delta_r^{(z)} \sim U(0.5, 1)$ is a random scaling factor for link $r$ in scenario $z$, drawn from a uniform distribution $U(0.5, 1)$. This perturbation process results in a reduction of the capacity of each link by up to 50\%, yielding various degrees of infrastructure degradation across the network.

\paragraph{Uniform Traffic Demand Perturbation (Dataset A: Random Uniform Scaling)}
The first dataset, hereafter referred to as Dataset A, leverages random uniform scaling to represent typical daily variations in traffic demand. The demand matrix for each scenario $z$ is generated by independently scaling the original demand $d_{ij}$ associated with each \gls{od} pair in the original demand matrix $\mathbf{D}$ as follows:
\begin{equation} \label{eq:demand-generation}
  d_{ij}^{(z)} = d_{ij} \delta_{ij}^{(z)}
\end{equation}
where $d_{ij}^{(z)}$ is the perturbed demand for the \gls{od} pair $(i,j)$ in scenario $z$, and $\delta_{ij}^{(z)} \sim U(0.5, 1.5)$ is its associated random scaling factor drawn from a uniform distribution $U(0.5, 1.5)$.

This process allows for both increases and decreases in demand for each \gls{od} pair, simulating typical daily fluctuations in traffic patterns while maintaining the overall spatial distribution of demand across the network.

\paragraph{Dirichlet Traffic Demand Perturbation (Dataset B: Distribution Shift)}
The second dataset, hereafter referred to as Dataset B, introduces a distribution shift to evaluate the generalization capabilities of the model under more extreme and structurally novel demand conditions, as its demand generation process fundamentally departs from the baseline spatial distribution of the original demand matrix $\mathbf{D}$. The primary objective of this dataset is to assess whether the \gls{gnn} can internalize the underlying physical principles of the \gls{tap}, i.e. flow conservation and Wardrop's \gls{ue}, rather than learning historical spatial correlations present in the training data.

To achieve this, every scenario in the dataset is structurally decoupled from the original demand matrix $\mathbf{D}$. Instead of applying localized scaling factors, the total network demand $D_{tot} = \sum_{i,j} d_{ij}$ is redistributed across all possible \gls{od} pairs using a probability simplex generated by a symmetric Dirichlet distribution \citep{bishop2006pattern}. For a given scenario $z$, the new demand entries are computed as:
\begin{equation} \label{eq:dirichlet-generation}
  d_{ij}^{(z)} = D_{tot} \cdot p_{ij}^{(z)}
\end{equation}
where $p_{ij}^{(z)}$ is the element of the probability vector $\mathbf{p}^{(z)}$ corresponding to the \gls{od} pair $(i,j)$, drawn from a symmetric Dirichlet distribution with concentration parameter $\alpha$:
\begin{equation}
  \mathbf{p}^{(z)} \sim Dir(\alpha \cdot \mathbf{1}_K)
\end{equation}
where $K$ is the total number of \gls{od} pairs in the network, and $\mathbf{1}_K$ is a $K$-dimensional vector of ones.

The Dirichlet distribution is a multivariate generalization of the Beta distribution, commonly used to generate random probability vectors that sum to 1 \citep{bishop2006pattern}. By adjusting the concentration parameter $\alpha$, we can control the sparsity and skewness of the generated demand patterns, with lower values ($\alpha \ll 1$) producing highly sparse and heavy-tailed distributions. A concentration parameter of $\alpha = 0.05$ was empirically determined to produce a demand distribution that closely resembled the original demand matrix in terms of sparsity and skewness, while being fundamentally different from the original baseline in terms of spatial distribution.

It is of fundamental importance to note that, as the resulting spatial distributions of the trips are mathematically randomized and do not follow realistic geographic constraints, Dataset B serves strictly as a synthetic stress-test benchmark. This design ensures that the model is evaluated purely on its robustness and its capacity to internalize the underlying physics of traffic flow, rather than relying on plausible, historical urban mobility patterns for its predictions.

\paragraph{Ground Truth Generation: \gls{ue} Solutions}
The final step in the dataset generation process is to compute the ground-truth \gls{ue} flow solutions for each generated scenario. The macroscopic assignments for each scenario were computed by solving the deterministic \gls{tap} using the \gls{bfw} algorithm implemented in the \textit{AequilibraE} framework \citep{camargo2015aequilibrae}. To ensure that the computed flow solutions represent a mathematically stable equilibrium state, the convergence criterion for the \gls{bfw} algorithm was set to a rgap threshold of $10^{-5}$, as suggested by \cite{boyce2004convergence}.

\subsubsection{Comparison of Uniform Scaling (Dataset A) and Distribution Shift (Dataset B)}
\begin{figure}[hbt]
  \centering
  \includegraphics[width=\textwidth]{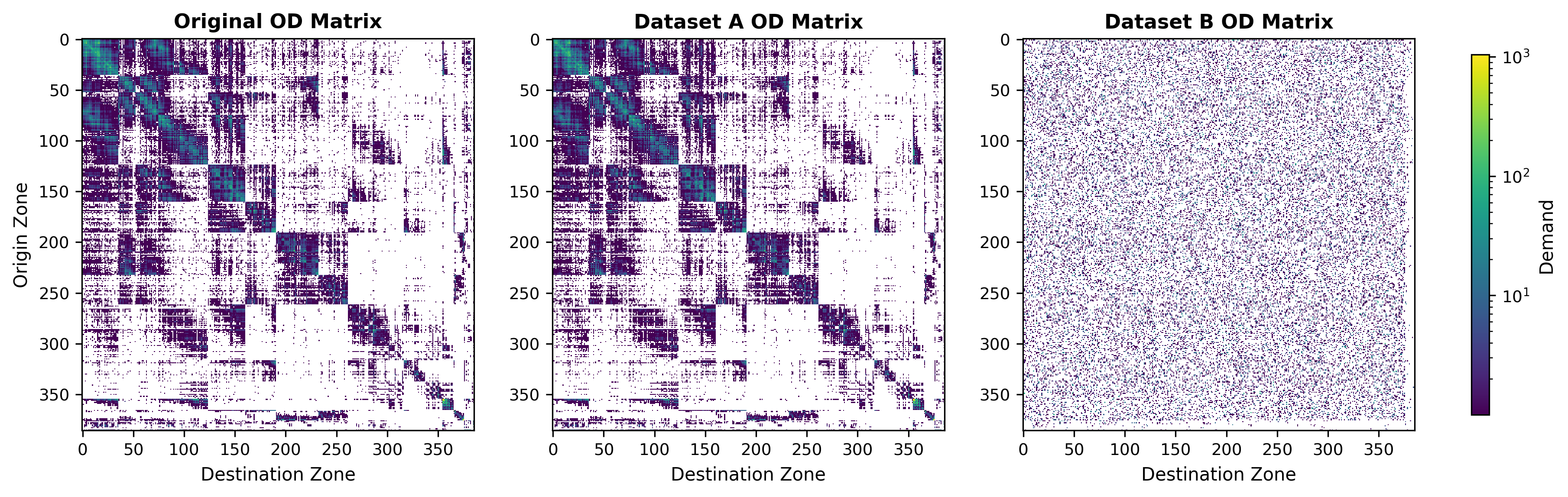}
  \captionof{figure}{Comparison of the spatial distribution of the original demand matrix for the Anaheim network (left) with two random synthesized demand matrices from Dataset A (Random Uniform Scaling, center) and B (Distribution Shift, right).}
  \label{fig:od-matrix-comparison}
\end{figure}

To visually illustrate the differences between the two datasets, \cref{fig:od-matrix-comparison} contrasts the spatial distribution of the baseline Anaheim demand matrix with two random synthesized demand matrices from Dataset A and B. While the demand matrix generated by the uniform scaling of Dataset A maintains a similar spatial structure to the original demand, the distribution shift of Dataset B leads to a demand matrix that exhibits a completely different spatial distribution, creating new demand patterns that are not correlated with the original demand matrix.

\begin{figure}[hbt]
  \centering
  \includegraphics[width=\textwidth]{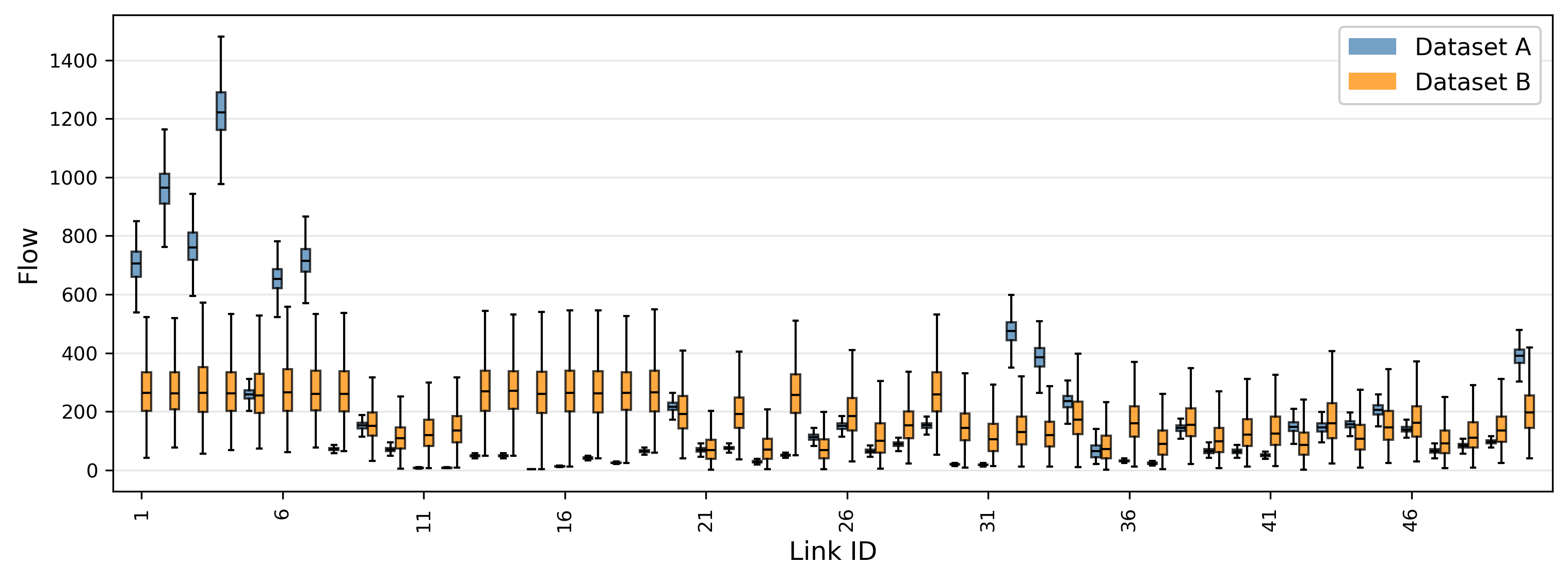}
  \captionof{figure}{Comparison of the distribution of link capacities and equilibrium flows across scenarios in Datasets A (Random Uniform Scaling) and B (Distribution Shift), for a subset of 50 links in the Anaheim network.}
  \label{fig:link-capacity-flow-box}
\end{figure}

Furthermore, \cref{fig:link-capacity-flow-box} compares the distribution of link equilibrium flows across all scenarios for 50 randomly selected links. As both datasets share an identical capacity perturbation process, any shift in the resulting traffic assignment is driven solely by the differences in the demand generation process. The distribution shift synthesized in Dataset B manifests as significantly higher flow variability and a wider range of extremes across all scenarios. Conversely, the uniform, daily variations generated in Dataset A lead to flow volumes that remain concentrated around the original baseline values for most links, with a much narrower variability across scenarios. This confirms that Dataset B is designed to be more challenging for the \gls{gnn} model, as it requires the model to learn the underlying physical principles of the \gls{tap} rather than relying on historical spatial correlations present in the training data.

\subsection{Training Framework} \label{sec:training-procedure}
Following the definition of the datasets, this section describes the preprocessing, optimization strategy, and implementation details used to train both the baseline \gls{hetgat} and the proposed \gls{model} models.

\subsubsection{Feature Engineering} \label{sec:feature-engineering}
All continuous input features were normalized independently using Min-Max scaling. Physical edge attributes, such as link capacities and free-flow travel times, were normalized to a $[0, 1]$ range, while the \gls{od} demands were scaled to a $[0, 100]$ range. The prediction target for the regression task was defined as the \gls{vcr}, rather than the raw flow values, as suggested by \cite{liu2024end}, and normalized to a range of $[0, 1]$ using Min-Max scaling.

To prevent data leakage, normalization parameters were computed strictly from the training subset of each dataset, utilizing a standard 70\%-20\%-10\% split for training, validation, and testing, respectively. This ensures that the normalization parameters are derived solely from the training data. During inference, the predicted \gls{vcr} values are inverse-transformed back to their original scale before evaluating the model's performance.

For the baseline \gls{hetgat} implementation, node features are obtained directly from the \gls{od} matrix, and therefore depend on the size of the underlying transportation network (\glspl{taz}). To enable experiments involving multiple network topologies, the baseline node feature vectors were zero-padded to a common dimensionality of 950 nodes. This dimension includes a small buffer over the largest considered network (Chicago) to ensure flexibility for potential minor topological modifications without requiring other architectural changes. This homogenization was applied exclusively to the baseline implementation; in contrast, the proposed \gls{model} models do not require any input dimensionality adjustments as they rely on the network-agnostic \gls{layer} initialization process.

\subsubsection{Optimization Strategy and Implementation Details}
The models were optimized using the Adam optimizer \citep{kingma2014adam} with an initial learning rate of $10^{-3}$. To refine the weights near convergence, we also applied a LR scheduler based on validation set performance, halving the learning rate as validation improvements stalled.

The maximum number of epochs was set to 700 for intra-network tasks, and 400 for inter-network transfer learning and fine-tuning experiments. The batch size was set to 16 for intra-network tasks, and 8 for inter-network and reduced size dataset tasks. Early Stopping was enabled across all runs, monitoring the validation loss to prevent overfitting and ensure the preservation of the most generalizable model weights.

To balance the gradient magnitudes during backpropagation, the weights for the composite loss function (\cref{sec:loss-function}) components were empirically set to $\lambda_v = 1.0$, $\lambda_f = 0.003$, and $\lambda_c = 0.003$ after a preliminary manual tuning phase. This scaling is mathematically necessary because the target variables operate in different domains: the \gls{vcr} predictions are bounded within a small range (as the target is normalized to $[0, 1]$), while the flow predictions and node conservation violations can span a much wider range of values, depending on the link capacities and demand levels.

Early testing showed that the simultaneous optimization of the purely supervised \gls{vcr} loss and the stringent flow conservation constraint from the very first epoch often led to unstable training dynamics, particularly in the early phases when the model's predictions are still far from accurate. To mitigate this issue, we implemented an epoch-based curriculum learning strategy, where the conservation loss is gradually introduced during training. This allows the model to first learn to approximate the \gls{vcr} values without being penalized for flow conservation violations, and then progressively encourages the model to internalize the physical constraint as its predictions become more accurate.

Both models were implemented in Python using the PyTorch and PyTorch Geometric libraries, which provide efficient implementations of graph neural network layers and operations \citep{Ansel_PyTorch_2_Faster_2024,Fey_PyG_2_0_Scalable_2025}.
The computations were performed on a workstation equipped with an NVIDIA RTX 4090 GPU (24GB VRAM) and an Intel i7-13700K CPU.

\subsection{Evaluation Metrics} \label{sec:evaluation-metrics}
To comprehensively assess the predictive accuracy and physical consistency of the tested models, we evaluate both standard statistical regression metrics and domain-specific transportation metrics.

\subsubsection{Statistical Evaluation Metrics}
These metrics evaluate the point-wise and global regression performance of the neural network architectures, quantifying the magnitude and variance of the prediction errors. They are computed on the predicted flows $\hat{f}_r$ for the real edges $r \in \mathcal{E}_r$, compared against the ground-truth flows $f_r$ obtained from a traditional \gls{ue} solver.

The choice of the metrics is motivated by their widespread use in regression tasks and their ability to provide insights into different aspects of model performance: \gls{mae} captures the typical size of the errors, \gls{rmsn} accounts for the scale of the flows, and the coefficient of determination ($R^2$) assesses the overall goodness-of-fit of the model to the data. These metrics together provide a comprehensive picture of how well the model is able to predict traffic flows across the network from a purely statistical and regression perspective.

\subsubsection{Physical and Transportation Metrics}
Domain-specific metrics, as commonly used in traffic modeling and transportation engineering, are employed to validate the physical realism of the predicted routing patterns. These metrics assess not only the accuracy of the flow predictions but also their adherence to fundamental traffic principles and their practical acceptability for real-world applications.

Two main metrics are utilized in this context: the percentage of links with a GEH statistic less than 5, which serves as a practical engineering acceptability criterion \citep{dowling2004traffic}, and the \gls{fcn}, which evaluates the degree to which the predicted flows adhere to the principle of flow conservation at the network nodes \citep{liu2024end}.

The \gls{fcn} metric quantifies the conservation loss across all nodes in the graph, normalized by the total demand in the network. This metric is particularly important for assessing the physical consistency of the predicted flows, as it directly measures whether the model's predictions respect the fundamental principle that the total inflow and outflow at each node must balance with the net demand generated or attracted at that node. It is computed as follows:
\begin{equation}
  FCN = \frac{\sum_{i \in \mathcal{V}} \left| \sum_{r \in \mathcal{E}_r^{out,i}} \hat{f}_r - \sum_{r \in \mathcal{E}_r^{in,i}} \hat{f}_r - d_{net,i} \right|}{\sum_{i \in \mathcal{V}} |d_{net,i}|}
\end{equation}
where $\mathcal{E}_r^{out,i}$ and $\mathcal{E}_r^{in,i}$ denote the sets of outgoing and incoming real edges incident to node $i$, respectively, and $d_{net,i}$ represents the net generated or attracted demand at node $i$, while $\hat{f}_r$ and $f_r$ denote the model-predicted and ground-truth flows on physical link $r \in \mathcal{E}_r$, respectively. A lower FCN value indicates better adherence to flow conservation, with a value of zero indicating perfect conservation across the network.

\subsection{Design of Experiments} \label{sec:experiment-design}
To systematically evaluate the proposed \gls{gnn} architectures and their capacity to learn generalized traffic physics, the experimental framework was structured into four distinct phases. Each phase introduces increasing levels of complexity, transitioning from strictly controlled intra-network conditions to transfer learning across structurally disparate networks, and finally to semi-supervised inference under partial observability. This progression allows for a comprehensive assessment of the models' generalization capabilities, robustness to data scarcity, and adaptability to real-world deployment scenarios.

A high-level overview of the complete experimental framework, detailing the data setup and primary research objectives for each phase, is provided in \cref{tab:experiment-summary}. The datasets generated for the Anaheim and Chicago networks, as described in \cref{sec:dataset-definition}, serve as the basis for all experiments. The specific configurations and objectives of each experimental phase are detailed in the subsequent sections.

\begin{table}[!bp]
  \footnotesize
  \renewcommand{\arraystretch}{1.3}
  \begin{tabularx}{\textwidth}{@{} l >{\raggedright\arraybackslash}p{3cm} X X @{}}
    \toprule
    \textbf{Exp.} & \textbf{Focus} & \textbf{Data Setup} & \textbf{Primary Objective} \\
    \midrule
    \textbf{A} & Intra-Network Generalization & Full training set (700 scenarios) trained and evaluated on the same network topology. & Establish the performance ceiling and verify that the proposed model is strictly competitive with the state-of-the-art on standard tasks. \\

    \textbf{B} & Sample Efficiency & Severely restricted training set (200 scenarios) trained and evaluated on the same network topology. & Assess model robustness to data scarcity and evaluate if structural inductive biases prevent overfitting compared to the baseline. \\

    \textbf{C} & Semi-Supervised Inference & Partial observability: models trained and evaluated with a masked percentage of physical link flows. & Determine the models' capacity to reconstruct unobserved macroscopic network states and maintain flow conservation from sparse data. \\

    \textbf{D} & Inter-Network Transferability & Bi-directional transfer: trained on source network (full dataset), zero-shot and fine-tuned (200 scenarios) on target network. & Evaluate spatial generalizability across disparate topologies and isolate transferable physical priors via layer-wise ablation. \\

    \bottomrule
  \end{tabularx}
  \caption{Summary of the experimental framework, detailing the training setup and primary research objectives for each phase.}
  \label{tab:experiment-summary}
\end{table}

\subsubsection{Experiment A: Intra-Network Generalization}
The primary objective of this baseline experiment is to establish the performance ceiling of the models and verify that the proposed architecture remains strictly competitive with the state-of-the-art on standard tasks. Before assessing complex cross-network transferability, it is crucial to ensure that the introduction of the high-dimensional virtual edge initialization does not compromise the model's fundamental ability to accurately map demand to \gls{ue} flows within the same topological domain.

To achieve this, both the baseline \gls{hetgat} and the proposed \gls{model} models are trained from scratch on all datasets [\gls{an-a}, \gls{an-b}, \gls{ch-a}, \gls{ch-b}], and evaluated on unseen demand distributions from the same respective networks.

\subsubsection{Experiment B: Sample Efficiency under Data Scarcity}
Data availability is a critical concern in real-world applications of deep learning frameworks. Gathering high-quality traffic data or generating macroscopic \gls{ue} solutions for a large number of scenarios is computationally expensive and time-consuming. Therefore, this experiment tests the models' practical applicability under real-world data scarcity, and verifies whether the additional parameter complexity introduced by the proposed \gls{layer} initialization hinders its sample efficiency compared to the simpler baseline.

To assess this, the training data is severely restricted to a subset of only 200 scenarios for each dataset [\gls{an-a}, \gls{an-b}, \gls{ch-a}, \gls{ch-b}], resulting in exactly 140 training, 40 validation, and 20 testing samples following the established 70-10-20 split. By analyzing the degradation in evaluation metrics relative to the fully-trained models in Experiment A, this phase explicitly benchmarks whether the richer node embeddings of the proposed architecture provide a competitive advantage over the state-of-the-art under constrained training conditions, or if the increased parameter complexity leads to overfitting and reduced generalization when data is scarce.

\subsubsection{Experiment C: Semi-Supervised Inference with Partial Observability}
In real-world traffic management systems, it is often the case that ground-truth flow data is not available for every physical link in the network. Instead, traffic data is typically collected through a sparse network of sensors, such as induction loops or camera-based traffic monitoring systems, which only provide partial observability of the overall traffic state. This experiment simulates such a semi-supervised deployment scenario to evaluate the models' ability to reconstruct unobserved network states based on limited information. During training, a masking function is applied to the physical edges $\mathcal{E}_r$, hiding a defined percentage (50\%) of the ground-truth link flows.

The objective is to determine if the combination of the custom virtual-edge architecture and the physics-informed flow conservation loss ($\mathcal{L}_c$) is sufficient to accurately reconstruct the unobserved network states based solely on partial link information, and to quantify the performance advantage provided by the richer embeddings of the proposed \gls{layer} layer.

\subsubsection{Experiment D: Inter-Network Transfer Learning and Layer Sensitivity Analysis}
This phase constitutes the core contribution regarding spatial transferability. The objective is to evaluate whether a model trained on a specific urban topology can successfully map demand to \gls{ue} flows on a completely unseen, structurally disparate network. To rigorously assess this, the transfer learning evaluation is conducted bi-directionally: models are trained on the Anaheim datasets [\gls{an-a}, \gls{an-b}] and transferred to the Chicago topology, as well as trained on the Chicago datasets [\gls{ch-a}, \gls{ch-b}] and transferred to the Anaheim topology.

For all inter-network tasks, the baseline \gls{hetgat} model utilizes zero-padding to standardize the input dimensions across both networks, as established in \cref{sec:feature-engineering}. The pre-trained weights from the source network are transferred to the target topology, where the model's performance is evaluated under two paradigms: Zero-Shot Inference (direct prediction on the target network without any weight updates) and Fine-Tuning (retraining on a limited subset of 200 scenarios from the target domain's dataset).

To further dissect the internal representations learned by the models and isolate how they encode topological features versus generalized flow physics, a layer-wise sensitivity analysis is conducted during the fine-tuning phase. By selectively freezing and fine-tuning specific components of the \gls{gnn}, we aim to isolate which learned representations are responsible for adapting to the new graph features, and which encode the underlying flow physics that can be transferred across topologies. This analysis provides insights into the internal mechanisms of the models and identifies which architectural components to prioritize for transfer learning in future applications and real-world deployment.

\section{Experimental Results and Discussion} \label{ch:results}
This section presents the empirical evaluation of the proposed \gls{gnn} architectures for solving the macroscopic \gls{tap}. Following the experimental design formulated in \cref{ch:experiment-setup}, the performance of the baseline \gls{hetgat} is systematically compared against the \gls{layer}-Enhanced formulations (\gls{model}-lin and \gls{model}-rbf) across multiple urban network topologies.

The results for Experiments A, B, and C are presented in \cref{tab:exp-abc-results}, which summarizes the performance metrics for each model under the respective experimental conditions. Results for Experiment D are presented in \cref{tab:exp-c_dataset_a-results} and \cref{tab:exp-c_dataset_b-results}. The empirical results of each experiment and their interpretation are presented in tandem, with a critical discussion of the observed performance differences.

\subsection{Experiment A: Intra-Network Generalization} \label{sec:experiment-a}
The results of Experiment A, which evaluates the performance of the baseline and proposed \gls{gnn} models on Datasets A and B for the Anaheim and Chicago networks, are presented in the first section of \cref{tab:exp-abc-results}.

Under the random uniform scaling of Dataset A, the baseline model already demonstrates high predictive accuracy. However, both \gls{model} variants outperform it on all metrics on both networks, showing substantial improvements especially in terms of the \gls{fcn} metric (e.g. 0.935 vs 0.416 on \gls{an-a}, 0.607 vs 0.339 on \gls{ch-a} for \gls{model}-lin).

\begin{table}[hbtp]
  \scriptsize
  \renewcommand{\arraystretch}{0.85}
  \newcommand{\firstcellgray}[1]{\multicolumn{1}{>{\columncolor{gray!15}[2pt][\tabcolsep]}l@{}}{#1}}
  \newcommand{\lastcellgray}[1]{\multicolumn{1}{>{\columncolor{gray!15}[\tabcolsep][1pt]}r@{}}{#1}}
  \begin{tabularx}{\textwidth}{@{}p{2.1cm}llXrrrrr@{}}
    \toprule
    \textbf{Experiment} & \textbf{Dataset} & \textbf{Network} & \textbf{Model} & \textbf{\textdownarrow{} MAE} & \textbf{\textdownarrow{} RMSN} & \textbf{\textuparrow{} $R^2$} & \textbf{\textuparrow{} \% GEH $< 5$} & \textbf{\textdownarrow{} FCN} \\
    \midrule
    \multirow{12}{=}{Experiment A: \textit{Intra-network Generalization}} & \multirow{6}{*}{Dataset A} & \multirow{3}{*}{Anaheim} & HetGAT & 23.14 & 0.256 & 0.955 & 93.92 & 0.935 \\
    & & & \firstcellgray{\gls{model}-lin} & \cellcolor{gray!15}\textbf{16.61} & \cellcolor{gray!15}\textbf{0.138} & \cellcolor{gray!15}\textbf{0.987} & \cellcolor{gray!15}\textbf{95.78} & \lastcellgray{\textbf{0.416}} \\
    & & & \firstcellgray{\gls{model}-rbf} & \cellcolor{gray!15}16.74 & \cellcolor{gray!15}0.140 & \cellcolor{gray!15}0.986 & \cellcolor{gray!15}95.53 & \lastcellgray{\textbf{0.416}} \\
    \cmidrule(l){3-9}
    & & \multirow{3}{*}{Chicago} & HetGAT & 17.52 & 0.122 & 0.987 & 98.59 & 0.607 \\
    & & & \firstcellgray{\gls{model}-lin} & \cellcolor{gray!15}14.09 & \cellcolor{gray!15}0.104 & \cellcolor{gray!15}0.991 & \cellcolor{gray!15}98.77 & \lastcellgray{0.339} \\
    & & & \firstcellgray{\gls{model}-rbf} & \cellcolor{gray!15}\textbf{13.38} & \cellcolor{gray!15}\textbf{0.098} & \cellcolor{gray!15}\textbf{0.992} & \cellcolor{gray!15}\textbf{98.95} & \lastcellgray{\textbf{0.334}} \\
    \cmidrule(l){2-9}
    & \multirow{6}{*}{Dataset B} & \multirow{3}{*}{Anaheim} & HetGAT & 51.22 & 0.380 & 0.814 & 66.07 & 1.206 \\
    & & & \firstcellgray{\gls{model}-lin} & \cellcolor{gray!15}\textbf{38.81} & \cellcolor{gray!15}0.297 & \cellcolor{gray!15}\textbf{0.887} & \cellcolor{gray!15}75.95 & \lastcellgray{\textbf{0.882}} \\
    & & & \firstcellgray{\gls{model}-rbf} & \cellcolor{gray!15}38.97 & \cellcolor{gray!15}\textbf{0.294} & \cellcolor{gray!15}0.886 & \cellcolor{gray!15}\textbf{76.04} & \lastcellgray{0.893} \\
    \cmidrule(l){3-9}
    & & \multirow{3}{*}{Chicago} & HetGAT & 93.46 & 0.184 & 0.950 & 74.89 & 2.417 \\
    & & & \firstcellgray{\gls{model}-lin} & \cellcolor{gray!15}64.23 & \cellcolor{gray!15}\textbf{0.146} & \cellcolor{gray!15}\textbf{0.968} & \cellcolor{gray!15}87.09 & \lastcellgray{\textbf{1.251}} \\
    & & & \firstcellgray{\gls{model}-rbf} & \cellcolor{gray!15}\textbf{63.91} & \cellcolor{gray!15}\textbf{0.146} & \cellcolor{gray!15}\textbf{0.968} & \cellcolor{gray!15}\textbf{87.26} & \lastcellgray{1.252} \\
    \midrule
    \multirow{12}{=}{Experiment B: \textit{Sample Efficiency under Data Scarcity}} & \multirow{6}{*}{Dataset A} & \multirow{3}{*}{Anaheim} & HetGAT & 24.15 & 0.265 & 0.951 & 93.00 & 1.038 \\
    & & & \firstcellgray{\gls{model}-lin} & \cellcolor{gray!15}17.95 & \cellcolor{gray!15}0.154 & \cellcolor{gray!15}0.984 & \cellcolor{gray!15}\textbf{95.40} & \lastcellgray{0.619} \\
    & & & \firstcellgray{\gls{model}-rbf} & \cellcolor{gray!15}\textbf{17.85} & \cellcolor{gray!15}\textbf{0.148} & \cellcolor{gray!15}\textbf{0.985} & \cellcolor{gray!15}94.97 & \lastcellgray{\textbf{0.571}} \\
    \cmidrule(l){3-9}
    & & \multirow{3}{*}{Chicago} & HetGAT & 18.19 & 0.127 & 0.986 & \textbf{98.29} & 0.693 \\
    & & & \gls{model}-lin & 18.18 & 0.121 & \textbf{0.988} & 97.78 & 0.554 \\
    & & & \gls{model}-rbf & \textbf{16.35} & \textbf{0.117} & \textbf{0.988} & 98.12 & \textbf{0.543} \\
    \cmidrule(l){2-9}
    & \multirow{6}{*}{Dataset B} & \multirow{3}{*}{Anaheim} & HetGAT & 50.71 & 0.368 & 0.823 & 66.91 & 1.679 \\
    & & & \firstcellgray{\gls{model}-lin} & \cellcolor{gray!15}\textbf{41.05} & \cellcolor{gray!15}\textbf{0.305} & \cellcolor{gray!15}\textbf{0.881} & \cellcolor{gray!15}\textbf{74.92} & \lastcellgray{1.503} \\
    & & & \firstcellgray{\gls{model}-rbf} & \cellcolor{gray!15}41.29 & \cellcolor{gray!15}\textbf{0.305} & \cellcolor{gray!15}0.875 & \cellcolor{gray!15}74.28 & \lastcellgray{\textbf{1.335}} \\
    \cmidrule(l){3-9}
    & & \multirow{3}{*}{Chicago} & HetGAT & 96.54 & 0.187 & 0.947 & 74.73 & 2.583 \\
    & & & \firstcellgray{\gls{model}-lin} & \cellcolor{gray!15}85.34 & \cellcolor{gray!15}0.183 & \cellcolor{gray!15}0.950 & \cellcolor{gray!15}79.98 & \lastcellgray{1.866} \\
    & & & \firstcellgray{\gls{model}-rbf} & \cellcolor{gray!15}\textbf{72.71} & \cellcolor{gray!15}\textbf{0.168} & \cellcolor{gray!15}\textbf{0.959} & \cellcolor{gray!15}\textbf{83.58} & \lastcellgray{\textbf{1.801}} \\
    \midrule
    \multirow{12}{=}{Experiment C: \textit{Semi-Supervised Learning under Partial Observability}} & \multirow{6}{*}{Dataset A} & \multirow{3}{*}{Anaheim} & HetGAT & 44.96 & 0.456 & 0.872 & 69.17 & 1.334 \\
    & & & \firstcellgray{\gls{model}-lin} & \cellcolor{gray!15}31.23 & \cellcolor{gray!15}0.250 & \cellcolor{gray!15}0.961 & \cellcolor{gray!15}75.73 & \lastcellgray{0.540} \\
    & & & \firstcellgray{\gls{model}-rbf} & \cellcolor{gray!15}\textbf{29.71} & \cellcolor{gray!15}\textbf{0.241} & \cellcolor{gray!15}\textbf{0.964} & \cellcolor{gray!15}\textbf{76.56} & \lastcellgray{\textbf{0.422}} \\
    \cmidrule(l){3-9}
    & & \multirow{3}{*}{Chicago} & HetGAT & 55.13 & 0.397 & 0.857 & 74.54 & 0.490 \\
    & & & \firstcellgray{\gls{model}-lin} & \cellcolor{gray!15}46.50 & \cellcolor{gray!15}0.360 & \cellcolor{gray!15}0.883 & \cellcolor{gray!15}78.91 & \lastcellgray{0.295} \\
    & & & \firstcellgray{\gls{model}-rbf} & \cellcolor{gray!15}\textbf{39.64} & \cellcolor{gray!15}\textbf{0.302} & \cellcolor{gray!15}\textbf{0.917} & \cellcolor{gray!15}\textbf{81.58} & \lastcellgray{\textbf{0.242}} \\
    \cmidrule(l){2-9}
    & \multirow{6}{*}{Dataset B} & \multirow{3}{*}{Anaheim} & HetGAT & 60.19 & 0.459 & 0.723 & 56.45 & 1.319 \\
    & & & \firstcellgray{\gls{model}-lin} & \cellcolor{gray!15}48.53 & \cellcolor{gray!15}0.380 & \cellcolor{gray!15}0.811 & \cellcolor{gray!15}65.08 & \lastcellgray{\textbf{0.790}} \\
    & & & \firstcellgray{\gls{model}-rbf} & \cellcolor{gray!15}\textbf{46.29} & \cellcolor{gray!15}\textbf{0.363} & \cellcolor{gray!15}\textbf{0.828} & \cellcolor{gray!15}\textbf{67.28} & \lastcellgray{0.826} \\
    \cmidrule(l){3-9}
    & & \multirow{3}{*}{Chicago} & HetGAT & 188.08 & 0.743 & 0.176 & 54.78 & 1.809 \\
    & & & \firstcellgray{\gls{model}-lin} & \cellcolor{gray!15}167.40 & \cellcolor{gray!15}0.435 & \cellcolor{gray!15}0.717 & \cellcolor{gray!15}62.29 & \lastcellgray{\textbf{1.693}} \\
    & & & \firstcellgray{\gls{model}-rbf} & \cellcolor{gray!15}\textbf{162.25} & \cellcolor{gray!15}\textbf{0.413} & \cellcolor{gray!15}\textbf{0.745} & \cellcolor{gray!15}\textbf{62.38} & \lastcellgray{1.724} \\
    \bottomrule
  \end{tabularx}%

  \scriptsize
  \vspace{1ex}
  Notes:\\Arrows indicate the desired direction of improvement for each metric; Dataset A refers to the Random Uniform Scaling Dataset; Dataset B refers to the Distribution Shift Dataset.\\Bold values denote the best overall performance for a specific metric within each combination.\\Highlighted rows indicate configurations where the proposed architecture outperforms the baseline across all evaluation metrics, for the specific Experiment / Dataset combination.
  \caption{Results and statistics from Experiments A, B, and C.}
  \label{tab:exp-abc-results}
\end{table}

On Dataset B, the performance gap between \gls{hetgat} and the proposed models widens significantly, with the baseline's predictive accuracy deteriorating substantially (e.g. $R^2$ drops from 0.955 to 0.814 on \gls{an-b}, and only 66\% of predictions fall within the GEH $< 5$ threshold). The proposed \gls{model} models, again, outperform it on all metrics on both networks, and maintain a much stronger performance, with an $R^2$ of around 0.886-0.887 and 76\% of predictions within the GEH threshold for \gls{an-b}. The \gls{fcn} metric also indicates a substantial enhancement in flow consistency, with the error decreasing from 1.206 to around 0.882-0.893 for \gls{an-b}.

\paragraph{Interpretation}
As both the baseline and proposed architectures operate on the same underlying heterogeneous graphs, the observed performance differences can be attributed to an enhanced representational capacity introduced by the rich, high-dimensional node features of the \gls{model} models.

The proposed \gls{layer} initialization process results in a more informative feature space, with node embeddings derived from a directional aggregation of the expanded input demand. Thus, the nodes inherently possess a holistic, localized representation of their role as trip generators or attractors, which is crucial for accurately modeling traffic flows and which is lost in the baseline \gls{hetgat}'s simpler initialization.

Overall, this richer feature space allows the \gls{model} models to learn more complex, non-linear relationships and routing behaviors between nodes, which is particularly beneficial under the more challenging conditions of the distribution shift in Dataset B. The consistent improvement across all metrics and datasets suggests that the proposed initialization strategy provides a more robust foundation for learning the underlying traffic flow dynamics, leading to superior predictive performance and better adherence to physical flow conservation principles.

\subsection{Experiment B: Sample Efficiency under Data Scarcity}
The results of Experiment B, which evaluates the performance of the baseline and proposed \gls{gnn} models on smaller subsets of Datasets A and B for the Anaheim and Chicago networks, are presented in the second section of \cref{tab:exp-abc-results}. The goal is to determine how robust the models are to data scarcity and whether the proposed high-dimensional initialization accelerates or hinders convergence under limited training samples.

Comparing these results to the fully trained models in Experiment A reveals distinct behavioral differences between the architectures. On Dataset A, all models still maintain high predictive accuracy, with only slight degradations in performance compared to their Experiment A counterparts. The baseline \gls{hetgat} exhibits surprising stability to the distribution shift of Dataset B: its performance on \gls{ch-b} remains largely unchanged, with its $R^2$ dropping only slightly from 0.950 to 0.947, and its GEH < 5 compliance maintaining a near-constant 74.73\% compared to its previous 74.89\%.

Conversely, the proposed \gls{model} architectures experience a noticeable degradation in performance compared to the Experiment A results. On \gls{ch-b}, the \gls{model}-lin model sees its \gls{mae} increase from 64.23 to 85.34, and its GEH < 5 compliance drops from 87.09\% to 79.98\%. However, despite this degradation, both \gls{model} models still comfortably outperform the baseline across all metrics on Dataset B for both networks.

\paragraph{Interpretation}
The contrasting responses to data scarcity highlight a fundamental trade-off between model capacity and sample efficiency. The baseline model's relatively static performance between Experiment A and B suggests that its simpler, transductive initialization limits its representational capacity; it effectively hits a performance ceiling that additional data (as in Experiment A) cannot overcome.

In contrast, the proposed \gls{layer} initialization introduces a richer, high-dimensional parameter space via the virtual edge expansions. While this increased capacity allows the \gls{model} models to achieve vastly superior accuracy when fully supplied with data (Experiment A), these additional parameters seem to be inherently more sensitive to data scarcity. When restricted to only 200 scenarios, they lack the samples required to fully optimize their expanded latent space, resulting in the observed performance drop. Crucially, however, the fact that their degraded performance still surpasses the baseline's performance ceiling confirms that the structural inductive bias provided by the virtual demand embeddings remains superior, allowing them to better generalize even under limited data conditions.

\subsection{Experiment C: Semi-Supervised Learning under Partial Observability}
The results of Experiment C, which evaluates the models' ability to reconstruct unobserved macroscopic network states from limited sensor data, are presented in the last section of \cref{tab:exp-abc-results}. In this semi-supervised deployment scenario, half of the physical link flows are masked during training: the objective is to determine if the models can leverage their respective demand representations and the physics-informed loss component to accurately deduce the unobserved traffic volumes. The reported metrics are then computed only on the unobserved subset of edges, which are the ones relevant for this specific task.

When compared to the performance of the models in \cref{sec:experiment-a}, the results of Experiment C reveal a significant performance degradation across all models and metrics, which is expected given the increased difficulty of the task.

While the proposed models still maintain moderate levels of predictive accuracy on the daily variations scenarios of Dataset A (e.g. GEH $< 5$ compliance of around 75\% on \gls{an-a} and around 80\% on \gls{ch-a}), the demand distribution shift of Dataset B leads to a significant performance drop for all models, with the baseline \gls{hetgat} struggling to provide acceptable predictions (e.g. GEH $< 5$ compliance of around 55\% on both networks). In contrast, the proposed \gls{model} models still manage to maintain a significantly better performance across all metrics, with GEH $< 5$ compliance rates of around 65\% on \gls{an-b} and around 62\% on \gls{ch-b}, which is a significant result given the difficulty of the task.

\paragraph{Interpretation}
The outcomes of this experiment, suggest once again that the proposed \gls{model} models, with their richer, high-dimensional demand representations, provide a more robust foundation for learning the underlying traffic flow dynamics, which is crucial for maintaining performance under conditions of partial observability.

Ultimately, these results, and especially the poor performance of both models on the challenging conditions of Dataset B, suggest that learning to approximate \gls{ue} flows from limited sensor data on large, dense networks with challenging demand patterns (such as those proposed by \gls{ch-b}) remains an open challenge. While the proposed \gls{model} architectures demonstrate significant improvements over the baseline model across all scenarios, the complex interactions between partial observability, network density, and high-dimensional demand representations warrant much deeper investigation. Extensive future testing, involving varying degrees of sensor sparsity, alternative edge-masking strategies, and diverse real-world topologies, is strictly necessary to fully investigate the potential and limits of the \gls{layer} layer integration in this specific field.

\subsection{Experiment D: Inter-Network Transfer Learning and Layer Sensitivity Analysis}
The results of Experiment D, which evaluates the transferability of the baseline and proposed \gls{gnn} models across different datasets and network topologies, are presented in \cref{tab:exp-c_dataset_a-results} and \cref{tab:exp-c_dataset_b-results}. The evaluation is structured as a layer sensitivity study, where different subsets of model parameters $\Theta$ are unfrozen for fine-tuning on the target network. This allows us to isolate the contributions of demand initialization versus topological routing components to the overall transferability of the models.

\begin{table}[htbp]
  \scriptsize
  \renewcommand{\arraystretch}{0.85}
  \begin{tabularx}{\textwidth}{@{}lXXrrrrr@{}}
    \toprule
    \textbf{Network} & \textbf{Model} & \textbf{Adapted Parameters} & \textbf{MAE \textdownarrow} & \textbf{RMSN \textdownarrow} & \textbf{$R^2$ \textuparrow} & \textbf{\% GEH $< 5$ \textuparrow} & \textbf{FCN \textdownarrow} \\ \midrule
    \multirow{33}{*}{Anaheim} & HetGAT & None & 309.39 & 1.933 & -2.201 & 21.77 & 6.922 \\
    & HetGAT & $\Theta_{pred}$ & 65.68 & 0.456 & 0.822 & 65.81 & 1.461 \\
    & HetGAT & $\Theta_{preproc}$ & 84.16 & 0.594 & 0.698 & 56.32 & 1.990 \\
    & HetGAT & $\Theta_{R_L}, \Theta_{pred}$ & 23.63 & 0.155 & 0.980 & 94.97 & 0.862 \\
    & HetGAT & $\Theta_{V_L}, \Theta_{pred}$ & 54.29 & 0.358 & 0.890 & 73.46 & 1.282 \\
    & HetGAT & $\Theta_{preproc}, \Theta_{pred}$ & 50.46 & 0.330 & 0.907 & 76.24 & 1.208 \\
    & HetGAT & $\Theta_{V_L}, \Theta_{R_L}, \Theta_{pred}$ & 22.30 & 0.147 & 0.982 & 96.03 & 0.819 \\
    & HetGAT & $\Theta_{preproc}, \Theta_{R_L}, \Theta_{pred}$ & 22.56 & 0.148 & 0.981 & 95.84 & 0.829 \\
    & HetGAT & $\Theta_{preproc}, \Theta_{V_L}, \Theta_{pred}$ & 43.21 & 0.281 & 0.933 & 81.80 & 1.172 \\
    & HetGAT & $\Theta_{preproc}, \Theta_{V_L}, \Theta_{R_L}, \Theta_{pred}$ & 21.39 & 0.141 & 0.983 & 96.95 & 0.816 \\
    & HetGAT & All & 19.58 & 0.134 & 0.985 & 97.70 & 0.735 \\
    & \gls{model}-lin & None & 170.82 & 1.106 & -0.050 & 26.33 & 6.924 \\
    & \cellcolor{gray!15} \gls{model}-lin & \cellcolor{gray!15} $\Theta_{pred}$ & \cellcolor{gray!15} 46.63 & \cellcolor{gray!15} 0.325 & \cellcolor{gray!15} 0.909 & \cellcolor{gray!15} 78.91 & \cellcolor{gray!15} 1.283 \\
    & \cellcolor{gray!15} \gls{model}-lin & \cellcolor{gray!15} $\Theta_{preproc}$ & \cellcolor{gray!15} 59.31 & \cellcolor{gray!15} 0.423 & \cellcolor{gray!15} 0.846 & \cellcolor{gray!15} 69.32 & \cellcolor{gray!15} 1.392 \\
    & \gls{model}-lin & $\Theta_{R_L}, \Theta_{pred}$ & 25.22 & 0.161 & 0.978 & 94.59 & 0.962 \\
    & \cellcolor{gray!15} \gls{model}-lin & \cellcolor{gray!15} $\Theta_{V_L}, \Theta_{pred}$ & \cellcolor{gray!15} 36.53 & \cellcolor{gray!15} 0.250 & \cellcolor{gray!15} 0.946 & \cellcolor{gray!15} 86.64 & \cellcolor{gray!15} 1.044 \\
    & \cellcolor{gray!15} \gls{model}-lin & \cellcolor{gray!15} $\Theta_{preproc}, \Theta_{pred}$ & \cellcolor{gray!15} 35.55 & \cellcolor{gray!15} 0.250 & \cellcolor{gray!15} 0.946 & \cellcolor{gray!15} 85.95 & \cellcolor{gray!15} 0.881 \\
    & \gls{model}-lin & $\Theta_{V_L}, \Theta_{R_L}, \Theta_{pred}$ & 23.30 & 0.149 & 0.981 & 95.91 & 0.851 \\
    & \cellcolor{gray!15} \gls{model}-lin & \cellcolor{gray!15} $\Theta_{preproc}, \Theta_{R_L}, \Theta_{pred}$ & \cellcolor{gray!15} 20.86 & \cellcolor{gray!15} 0.141 & \cellcolor{gray!15} 0.983 & \cellcolor{gray!15} 95.92 & \cellcolor{gray!15} 0.691 \\
    & \cellcolor{gray!15} \gls{model}-lin & \cellcolor{gray!15} $\Theta_{preproc}, \Theta_{V_L}, \Theta_{pred}$ & \cellcolor{gray!15} 27.43 & \cellcolor{gray!15} 0.188 & \cellcolor{gray!15} 0.970 & \cellcolor{gray!15} 91.56 & \cellcolor{gray!15} 0.783 \\
    & \gls{model}-lin & $\Theta_{preproc}, \Theta_{V_L}, \Theta_{R_L}, \Theta_{pred}$ & 20.73 & 0.138 & 0.984 & 96.24 & 0.685 \\
    & \cellcolor{gray!15} \gls{model}-lin & \cellcolor{gray!15} All & \cellcolor{gray!15} \textbf{16.59} & \cellcolor{gray!15} \textbf{0.116} & \cellcolor{gray!15} \textbf{0.988} & \cellcolor{gray!15} \textbf{98.25} & \cellcolor{gray!15} \textbf{0.528} \\
    & \gls{model}-rbf & None & 215.96 & 1.489 & -0.895 & 23.56 & 10.677 \\
    & \cellcolor{gray!15} \gls{model}-rbf & \cellcolor{gray!15} $\Theta_{pred}$ & \cellcolor{gray!15} 41.05 & \cellcolor{gray!15} 0.296 & \cellcolor{gray!15} 0.925 & \cellcolor{gray!15} 82.84 & \cellcolor{gray!15} 1.023 \\
    & \cellcolor{gray!15} \gls{model}-rbf & \cellcolor{gray!15} $\Theta_{preproc}$ & \cellcolor{gray!15} 57.06 & \cellcolor{gray!15} 0.402 & \cellcolor{gray!15} 0.862 & \cellcolor{gray!15} 70.59 & \cellcolor{gray!15} 1.281 \\
    & \cellcolor{gray!15} \gls{model}-rbf & \cellcolor{gray!15} $\Theta_{R_L}, \Theta_{pred}$ & \cellcolor{gray!15} 21.40 & \cellcolor{gray!15} 0.143 & \cellcolor{gray!15} 0.982 & \cellcolor{gray!15} 96.52 & \cellcolor{gray!15} 0.776 \\
    & \cellcolor{gray!15} \gls{model}-rbf & \cellcolor{gray!15} $\Theta_{V_L}, \Theta_{pred}$ & \cellcolor{gray!15} 32.10 & \cellcolor{gray!15} 0.227 & \cellcolor{gray!15} 0.956 & \cellcolor{gray!15} 89.62 & \cellcolor{gray!15} 0.818 \\
    & \cellcolor{gray!15} \gls{model}-rbf & \cellcolor{gray!15} $\Theta_{preproc}, \Theta_{pred}$ & \cellcolor{gray!15} 33.80 & \cellcolor{gray!15} 0.239 & \cellcolor{gray!15} 0.951 & \cellcolor{gray!15} 86.72 & \cellcolor{gray!15} 0.784 \\
    & \cellcolor{gray!15} \gls{model}-rbf & \cellcolor{gray!15} $\Theta_{V_L}, \Theta_{R_L}, \Theta_{pred}$ & \cellcolor{gray!15} 20.83 & \cellcolor{gray!15} 0.140 & \cellcolor{gray!15} 0.983 & \cellcolor{gray!15} 96.79 & \cellcolor{gray!15} 0.701 \\
    & \cellcolor{gray!15} \gls{model}-rbf & \cellcolor{gray!15} $\Theta_{preproc}, \Theta_{R_L}, \Theta_{pred}$ & \cellcolor{gray!15} 20.07 & \cellcolor{gray!15} 0.139 & \cellcolor{gray!15} 0.984 & \cellcolor{gray!15} 96.65 & \cellcolor{gray!15} 0.674 \\
    & \cellcolor{gray!15} \gls{model}-rbf & \cellcolor{gray!15} $\Theta_{preproc}, \Theta_{V_L}, \Theta_{pred}$ & \cellcolor{gray!15} 25.74 & \cellcolor{gray!15} 0.181 & \cellcolor{gray!15} 0.972 & \cellcolor{gray!15} 92.92 & \cellcolor{gray!15} 0.677 \\
    & \cellcolor{gray!15} \gls{model}-rbf & \cellcolor{gray!15} $\Theta_{preproc}, \Theta_{V_L}, \Theta_{R_L}, \Theta_{pred}$ & \cellcolor{gray!15} 18.47 & \cellcolor{gray!15} 0.129 & \cellcolor{gray!15} 0.986 & \cellcolor{gray!15} 97.42 & \cellcolor{gray!15} 0.592 \\
    & \cellcolor{gray!15} \gls{model}-rbf & \cellcolor{gray!15} All & \cellcolor{gray!15} 16.85 & \cellcolor{gray!15} 0.120 & \cellcolor{gray!15} 0.988 & \cellcolor{gray!15} 97.94 & \cellcolor{gray!15} 0.557 \\
    \cmidrule(l){1-8}
    \multirow{33}{*}{Chicago} & HetGAT & None & 162.54 & 1.313 & -0.203 & 23.74 & 11.305 \\
    & HetGAT & $\Theta_{pred}$ & 46.15 & 0.446 & 0.861 & 77.95 & 4.356 \\
    & HetGAT & $\Theta_{preproc}$ & 126.53 & 1.059 & 0.217 & 36.67 & 5.221 \\
    & HetGAT & $\Theta_{R_L}, \Theta_{pred}$ & 26.12 & 0.287 & 0.942 & 92.40 & 1.748 \\
    & HetGAT & $\Theta_{V_L}, \Theta_{pred}$ & 41.04 & 0.410 & 0.883 & 81.62 & 3.719 \\
    & HetGAT & $\Theta_{preproc}, \Theta_{pred}$ & 39.92 & 0.396 & 0.891 & 82.17 & 3.654 \\
    & HetGAT & $\Theta_{V_L}, \Theta_{R_L}, \Theta_{pred}$ & 26.02 & 0.288 & 0.942 & 92.47 & 1.881 \\
    & HetGAT & $\Theta_{preproc}, \Theta_{R_L}, \Theta_{pred}$ & 25.50 & 0.273 & 0.948 & 92.52 & 1.907 \\
    & HetGAT & $\Theta_{preproc}, \Theta_{V_L}, \Theta_{pred}$ & 39.95 & 0.392 & 0.893 & 82.04 & 3.657 \\
    & HetGAT & $\Theta_{preproc}, \Theta_{V_L}, \Theta_{R_L}, \Theta_{pred}$ & 25.29 & 0.270 & 0.949 & 92.78 & 1.778 \\
    & HetGAT & All & 23.83 & 0.264 & 0.951 & 93.70 & 1.360 \\
    & \gls{model}-lin & None & 171.73 & 1.276 & -0.130 & 22.08 & 13.919 \\
    & \cellcolor{gray!15} \gls{model}-lin & \cellcolor{gray!15} $\Theta_{pred}$ & \cellcolor{gray!15} 37.83 & \cellcolor{gray!15} 0.340 & \cellcolor{gray!15} 0.920 & \cellcolor{gray!15} 80.43 & \cellcolor{gray!15} 3.792 \\
    & \gls{model}-lin & $\Theta_{preproc}$ & 105.15 & 0.895 & 0.445 & 48.68 & 5.437 \\
    & \cellcolor{gray!15} \gls{model}-lin & \cellcolor{gray!15} $\Theta_{R_L}, \Theta_{pred}$ & \cellcolor{gray!15} 20.65 & \cellcolor{gray!15} 0.175 & \cellcolor{gray!15} 0.979 & \cellcolor{gray!15} 93.69 & \cellcolor{gray!15} 1.337 \\
    & \cellcolor{gray!15} \gls{model}-lin & \cellcolor{gray!15} $\Theta_{V_L}, \Theta_{pred}$ & \cellcolor{gray!15} 29.49 & \cellcolor{gray!15} 0.278 & \cellcolor{gray!15} 0.946 & \cellcolor{gray!15} 87.45 & \cellcolor{gray!15} 2.641 \\
    & \cellcolor{gray!15} \gls{model}-lin & \cellcolor{gray!15} $\Theta_{preproc}, \Theta_{pred}$ & \cellcolor{gray!15} 28.92 & \cellcolor{gray!15} 0.261 & \cellcolor{gray!15} 0.953 & \cellcolor{gray!15} 87.51 & \cellcolor{gray!15} 2.593 \\
    & \cellcolor{gray!15} \gls{model}-lin & \cellcolor{gray!15} $\Theta_{V_L}, \Theta_{R_L}, \Theta_{pred}$ & \cellcolor{gray!15} 19.80 & \cellcolor{gray!15} 0.170 & \cellcolor{gray!15} 0.980 & \cellcolor{gray!15} 94.06 & \cellcolor{gray!15} 1.151 \\
    & \cellcolor{gray!15} \gls{model}-lin & \cellcolor{gray!15} $\Theta_{preproc}, \Theta_{R_L}, \Theta_{pred}$ & \cellcolor{gray!15} 19.87 & \cellcolor{gray!15} 0.169 & \cellcolor{gray!15} 0.980 & \cellcolor{gray!15} 94.04 & \cellcolor{gray!15} 1.109 \\
    & \cellcolor{gray!15} \gls{model}-lin & \cellcolor{gray!15} $\Theta_{preproc}, \Theta_{V_L}, \Theta_{pred}$ & \cellcolor{gray!15} 27.77 & \cellcolor{gray!15} 0.254 & \cellcolor{gray!15} 0.955 & \cellcolor{gray!15} 88.56 & \cellcolor{gray!15} 2.408 \\
    & \cellcolor{gray!15} \gls{model}-lin & \cellcolor{gray!15} $\Theta_{preproc}, \Theta_{V_L}, \Theta_{R_L}, \Theta_{pred}$ & \cellcolor{gray!15} 19.66 & \cellcolor{gray!15} 0.172 & \cellcolor{gray!15} 0.980 & \cellcolor{gray!15} 94.14 & \cellcolor{gray!15} 0.984 \\
    & \cellcolor{gray!15} \gls{model}-lin & \cellcolor{gray!15} All & \cellcolor{gray!15} 17.96 & \cellcolor{gray!15} \textbf{0.149} & \cellcolor{gray!15} \textbf{0.985} & \cellcolor{gray!15} 95.03 & \cellcolor{gray!15} \textbf{0.688} \\
    & \gls{model}-rbf & None & 165.66 & 1.333 & -0.239 & 26.70 & 14.374 \\
    & \gls{model}-rbf & $\Theta_{pred}$ & 44.17 & 0.389 & 0.894 & 75.41 & 4.634 \\
    & \cellcolor{gray!15} \gls{model}-rbf & \cellcolor{gray!15} $\Theta_{preproc}$ & \cellcolor{gray!15} 103.39 & \cellcolor{gray!15} 0.891 & \cellcolor{gray!15} 0.446 & \cellcolor{gray!15} 47.55 & \cellcolor{gray!15} 4.906 \\
    & \cellcolor{gray!15} \gls{model}-rbf & \cellcolor{gray!15} $\Theta_{R_L}, \Theta_{pred}$ & \cellcolor{gray!15} 20.45 & \cellcolor{gray!15} 0.176 & \cellcolor{gray!15} 0.978 & \cellcolor{gray!15} 93.97 & \cellcolor{gray!15} 1.566 \\
    & \cellcolor{gray!15} \gls{model}-rbf & \cellcolor{gray!15} $\Theta_{V_L}, \Theta_{pred}$ & \cellcolor{gray!15} 30.66 & \cellcolor{gray!15} 0.280 & \cellcolor{gray!15} 0.945 & \cellcolor{gray!15} 86.21 & \cellcolor{gray!15} 2.790 \\
    & \cellcolor{gray!15} \gls{model}-rbf & \cellcolor{gray!15} $\Theta_{preproc}, \Theta_{pred}$ & \cellcolor{gray!15} 28.46 & \cellcolor{gray!15} 0.261 & \cellcolor{gray!15} 0.952 & \cellcolor{gray!15} 87.56 & \cellcolor{gray!15} 2.549 \\
    & \cellcolor{gray!15} \gls{model}-rbf & \cellcolor{gray!15} $\Theta_{V_L}, \Theta_{R_L}, \Theta_{pred}$ & \cellcolor{gray!15} 18.55 & \cellcolor{gray!15} 0.157 & \cellcolor{gray!15} 0.983 & \cellcolor{gray!15} 95.34 & \cellcolor{gray!15} 1.150 \\
    & \cellcolor{gray!15} \gls{model}-rbf & \cellcolor{gray!15} $\Theta_{preproc}, \Theta_{R_L}, \Theta_{pred}$ & \cellcolor{gray!15} 18.75 & \cellcolor{gray!15} 0.159 & \cellcolor{gray!15} 0.982 & \cellcolor{gray!15} 95.11 & \cellcolor{gray!15} 1.175 \\
    & \cellcolor{gray!15} \gls{model}-rbf & \cellcolor{gray!15} $\Theta_{preproc}, \Theta_{V_L}, \Theta_{pred}$ & \cellcolor{gray!15} 29.04 & \cellcolor{gray!15} 0.271 & \cellcolor{gray!15} 0.949 & \cellcolor{gray!15} 87.19 & \cellcolor{gray!15} 2.673 \\
    & \cellcolor{gray!15} \gls{model}-rbf & \cellcolor{gray!15} $\Theta_{preproc}, \Theta_{V_L}, \Theta_{R_L}, \Theta_{pred}$ & \cellcolor{gray!15} 18.77 & \cellcolor{gray!15} 0.161 & \cellcolor{gray!15} 0.982 & \cellcolor{gray!15} 95.10 & \cellcolor{gray!15} 1.182 \\
    & \cellcolor{gray!15} \gls{model}-rbf & \cellcolor{gray!15} All & \cellcolor{gray!15} \textbf{17.31} & \cellcolor{gray!15} 0.149 & \cellcolor{gray!15} 0.984 & \cellcolor{gray!15} \textbf{95.79} & \cellcolor{gray!15} 0.788 \\
    \bottomrule
  \end{tabularx}%

  \scriptsize
  \vspace{1ex}
  Notes:\\Arrows indicate the desired direction of improvement for each metric.\\Bold values denote the best overall performance for a specific metric within each combination.\\Highlighted rows indicate configurations where the proposed architecture outperforms the baseline across all evaluation metrics, for the specific Experiment / Dataset combination.
  \caption{Results and statistics from Experiment D (Dataset A, Random Uniform Scaling). Following a catastrophic failure in zero-shot transfer for all models, the \gls{model} architectures consistently outperform the baseline across most fine-tuning configurations.}
  \label{tab:exp-c_dataset_a-results}
\end{table}

\begin{table}[htbp]
  \scriptsize
  \renewcommand{\arraystretch}{0.85}
  \begin{tabularx}{\textwidth}{@{}lXXrrrrr@{}}
    \toprule
    \textbf{Network} & \textbf{Model} & \textbf{Adapted Parameters} & \textbf{\textdownarrow{} MAE} & \textbf{\textdownarrow{} RMSN} & \textbf{\textuparrow{} $R^2$} & \textbf{\textuparrow{} \% GEH $< 5$} & \textbf{\textdownarrow{} FCN} \\ \midrule
    \multirow{33}{*}{Anaheim} & HetGAT & None & 608.72 & 1.070 & -0.721 & 16.69 & 8.122 \\
    & HetGAT & $\Theta_{pred}$ & 189.83 & 0.393 & 0.768 & 51.01 & 3.506 \\
    & HetGAT & $\Theta_{preproc}$ & 244.47 & 0.550 & 0.546 & 43.61 & 4.809 \\
    & HetGAT & $\Theta_{R_L}, \Theta_{pred}$ & 126.60 & 0.244 & 0.911 & 63.13 & 2.592 \\
    & HetGAT & $\Theta_{V_L}, \Theta_{pred}$ & 181.76 & 0.377 & 0.787 & 52.94 & 3.327 \\
    & HetGAT & $\Theta_{preproc}, \Theta_{pred}$ & 166.01 & 0.336 & 0.830 & 55.36 & 3.952 \\
    & HetGAT & $\Theta_{V_L}, \Theta_{R_L}, \Theta_{pred}$ & 113.70 & 0.218 & 0.929 & 67.77 & 2.613 \\
    & HetGAT & $\Theta_{preproc}, \Theta_{R_L}, \Theta_{pred}$ & 110.54 & 0.213 & 0.932 & 69.04 & 2.792 \\
    & HetGAT & $\Theta_{preproc}, \Theta_{V_L}, \Theta_{pred}$ & 168.35 & 0.344 & 0.822 & 55.10 & 3.584 \\
    & HetGAT & $\Theta_{preproc}, \Theta_{V_L}, \Theta_{R_L}, \Theta_{pred}$ & 104.89 & 0.200 & 0.940 & 71.08 & 2.604 \\
    & HetGAT & All & 97.38 & 0.188 & 0.947 & 74.30 & 2.689 \\
    & \gls{model}-lin & None & 957.59 & 2.194 & -6.264 & 23.70 & 41.739 \\
    & \gls{model}-lin & $\Theta_{pred}$ & 181.58 & 0.405 & 0.752 & 58.13 & 2.835 \\
    & \cellcolor{gray!15} \gls{model}-lin & \cellcolor{gray!15} $\Theta_{preproc}$ & \cellcolor{gray!15} 223.88 & \cellcolor{gray!15} 0.522 & \cellcolor{gray!15} 0.589 & \cellcolor{gray!15} 51.63 & \cellcolor{gray!15} 4.196 \\
    & \cellcolor{gray!15} \gls{model}-lin & \cellcolor{gray!15} $\Theta_{R_L}, \Theta_{pred}$ & \cellcolor{gray!15} 104.69 & \cellcolor{gray!15} 0.226 & \cellcolor{gray!15} 0.923 & \cellcolor{gray!15} 73.31 & \cellcolor{gray!15} 2.284 \\
    & \gls{model}-lin & $\Theta_{V_L}, \Theta_{pred}$ & 169.01 & 0.377 & 0.785 & 60.13 & 2.677 \\
    & \gls{model}-lin & $\Theta_{preproc}, \Theta_{pred}$ & 167.84 & 0.370 & 0.793 & 60.03 & 2.457 \\
    & \cellcolor{gray!15} \gls{model}-lin & \cellcolor{gray!15} $\Theta_{V_L}, \Theta_{R_L}, \Theta_{pred}$ & \cellcolor{gray!15} 93.92 & \cellcolor{gray!15} 0.197 & \cellcolor{gray!15} 0.942 & \cellcolor{gray!15} 76.79 & \cellcolor{gray!15} 2.274 \\
    & \cellcolor{gray!15} \gls{model}-lin & \cellcolor{gray!15} $\Theta_{preproc}, \Theta_{R_L}, \Theta_{pred}$ & \cellcolor{gray!15} 96.35 & \cellcolor{gray!15} 0.207 & \cellcolor{gray!15} 0.935 & \cellcolor{gray!15} 75.95 & \cellcolor{gray!15} 2.196 \\
    & \gls{model}-lin & $\Theta_{preproc}, \Theta_{V_L}, \Theta_{pred}$ & 164.26 & 0.366 & 0.798 & 61.13 & 2.434 \\
    & \gls{model}-lin & $\Theta_{preproc}, \Theta_{V_L}, \Theta_{R_L}, \Theta_{pred}$ & 93.71 & 0.200 & 0.939 & 76.56 & 2.216 \\
    & \cellcolor{gray!15} \gls{model}-lin & \cellcolor{gray!15} All & \cellcolor{gray!15} \textbf{79.49} & \cellcolor{gray!15} \textbf{0.171} & \cellcolor{gray!15} \textbf{0.956} & \cellcolor{gray!15} \textbf{82.23} & \cellcolor{gray!15} 1.986 \\
    & \gls{model}-rbf & None & 1110.24 & 2.494 & -8.358 & 22.48 & 36.777 \\
    & \gls{model}-rbf & $\Theta_{pred}$ & 185.64 & 0.413 & 0.744 & 57.55 & 3.022 \\
    & \cellcolor{gray!15} \gls{model}-rbf & \cellcolor{gray!15} $\Theta_{preproc}$ & \cellcolor{gray!15} 228.14 & \cellcolor{gray!15} 0.518 & \cellcolor{gray!15} 0.596 & \cellcolor{gray!15} 50.64 & \cellcolor{gray!15} 4.390 \\
    & \cellcolor{gray!15} \gls{model}-rbf & \cellcolor{gray!15} $\Theta_{R_L}, \Theta_{pred}$ & \cellcolor{gray!15} 104.74 & \cellcolor{gray!15} 0.219 & \cellcolor{gray!15} 0.928 & \cellcolor{gray!15} 72.94 & \cellcolor{gray!15} 2.501 \\
    & \gls{model}-rbf & $\Theta_{V_L}, \Theta_{pred}$ & 174.70 & 0.383 & 0.779 & 59.13 & 2.695 \\
    & \gls{model}-rbf & $\Theta_{preproc}, \Theta_{pred}$ & 171.92 & 0.380 & 0.783 & 59.58 & 2.553 \\
    & \cellcolor{gray!15} \gls{model}-rbf & \cellcolor{gray!15} $\Theta_{V_L}, \Theta_{R_L}, \Theta_{pred}$ & \cellcolor{gray!15} 92.50 & \cellcolor{gray!15} 0.192 & \cellcolor{gray!15} 0.944 & \cellcolor{gray!15} 76.61 & \cellcolor{gray!15} 2.314 \\
    & \cellcolor{gray!15} \gls{model}-rbf & \cellcolor{gray!15} $\Theta_{preproc}, \Theta_{R_L}, \Theta_{pred}$ & \cellcolor{gray!15} 96.08 & \cellcolor{gray!15} 0.202 & \cellcolor{gray!15} 0.938 & \cellcolor{gray!15} 75.72 & \cellcolor{gray!15} 2.143 \\
    & \gls{model}-rbf & $\Theta_{preproc}, \Theta_{V_L}, \Theta_{pred}$ & 171.64 & 0.379 & 0.784 & 59.56 & 2.453 \\
    & \gls{model}-rbf & $\Theta_{preproc}, \Theta_{V_L}, \Theta_{R_L}, \Theta_{pred}$ & 94.83 & 0.201 & 0.939 & 76.51 & 2.125 \\
    & \cellcolor{gray!15} \gls{model}-rbf & \cellcolor{gray!15} All & \cellcolor{gray!15} 81.49 & \cellcolor{gray!15} 0.174 & \cellcolor{gray!15} 0.954 & \cellcolor{gray!15} 81.43 & \cellcolor{gray!15} \textbf{1.943} \\
    \cmidrule(l){1-8}
    \multirow{33}{*}{Chicago} & HetGAT & None & 164.29 & 1.226 & -0.950 & 16.94 & 6.554 \\
    & HetGAT & $\Theta_{pred}$ & 65.72 & 0.488 & 0.691 & 54.03 & 4.065 \\
    & HetGAT & $\Theta_{preproc}$ & 114.56 & 0.883 & -0.011 & 33.09 & 4.616 \\
    & HetGAT & $\Theta_{R_L}, \Theta_{pred}$ & 54.26 & 0.402 & 0.791 & 63.01 & 2.405 \\
    & HetGAT & $\Theta_{V_L}, \Theta_{pred}$ & 62.54 & 0.462 & 0.723 & 56.27 & 3.407 \\
    & HetGAT & $\Theta_{preproc}, \Theta_{pred}$ & 61.90 & 0.458 & 0.728 & 56.96 & 3.615 \\
    & HetGAT & $\Theta_{V_L}, \Theta_{R_L}, \Theta_{pred}$ & 53.67 & 0.397 & 0.795 & 63.59 & 2.078 \\
    & HetGAT & $\Theta_{preproc}, \Theta_{R_L}, \Theta_{pred}$ & 53.73 & 0.397 & 0.796 & 63.78 & 2.049 \\
    & HetGAT & $\Theta_{preproc}, \Theta_{V_L}, \Theta_{pred}$ & 62.05 & 0.459 & 0.726 & 56.94 & 3.322 \\
    & HetGAT & $\Theta_{preproc}, \Theta_{V_L}, \Theta_{R_L}, \Theta_{pred}$ & 53.66 & 0.397 & 0.796 & 63.61 & 2.078 \\
    & HetGAT & All & 52.61 & 0.390 & 0.802 & 64.44 & 1.877 \\
    & \gls{model}-lin & None & 181.45 & 1.315 & -1.232 & 10.81 & 1.853 \\
    & \gls{model}-lin & $\Theta_{pred}$ & 68.56 & 0.535 & 0.631 & 54.71 & 4.435 \\
    & \gls{model}-lin & $\Theta_{preproc}$ & 103.88 & 0.827 & 0.118 & 39.89 & 4.870 \\
    & \gls{model}-lin & $\Theta_{R_L}, \Theta_{pred}$ & 43.83 & 0.329 & 0.860 & 71.21 & 2.449 \\
    & \gls{model}-lin & $\Theta_{V_L}, \Theta_{pred}$ & 56.12 & 0.427 & 0.764 & 62.00 & 3.539 \\
    & \gls{model}-lin & $\Theta_{preproc}, \Theta_{pred}$ & 56.98 & 0.431 & 0.760 & 60.63 & 3.652 \\
    & \cellcolor{gray!15} \gls{model}-lin & \cellcolor{gray!15} $\Theta_{V_L}, \Theta_{R_L}, \Theta_{pred}$ & \cellcolor{gray!15} 42.13 & \cellcolor{gray!15} 0.319 & \cellcolor{gray!15} 0.869 & \cellcolor{gray!15} 73.23 & \cellcolor{gray!15} 2.001 \\
    & \cellcolor{gray!15} \gls{model}-lin & \cellcolor{gray!15} $\Theta_{preproc}, \Theta_{R_L}, \Theta_{pred}$ & \cellcolor{gray!15} 41.99 & \cellcolor{gray!15} 0.319 & \cellcolor{gray!15} 0.869 & \cellcolor{gray!15} 72.55 & \cellcolor{gray!15} 1.872 \\
    & \cellcolor{gray!15} \gls{model}-lin & \cellcolor{gray!15} $\Theta_{preproc}, \Theta_{V_L}, \Theta_{pred}$ & \cellcolor{gray!15} 53.90 & \cellcolor{gray!15} 0.408 & \cellcolor{gray!15} 0.786 & \cellcolor{gray!15} 63.12 & \cellcolor{gray!15} 3.245 \\
    & \cellcolor{gray!15} \gls{model}-lin & \cellcolor{gray!15} $\Theta_{preproc}, \Theta_{V_L}, \Theta_{R_L}, \Theta_{pred}$ & \cellcolor{gray!15} 41.79 & \cellcolor{gray!15} 0.316 & \cellcolor{gray!15} 0.871 & \cellcolor{gray!15} 73.06 & \cellcolor{gray!15} 1.942 \\
    & \cellcolor{gray!15} \gls{model}-lin & \cellcolor{gray!15} All & \cellcolor{gray!15} \textbf{39.60} & \cellcolor{gray!15} \textbf{0.301} & \cellcolor{gray!15} \textbf{0.883} & \cellcolor{gray!15} \textbf{75.66} & \cellcolor{gray!15} 1.682 \\
    & \gls{model}-rbf & None & 185.16 & 1.271 & -1.163 & 13.00 & 3.710 \\
    & \cellcolor{gray!15} \gls{model}-rbf & \cellcolor{gray!15} $\Theta_{pred}$ & \cellcolor{gray!15} 64.60 & \cellcolor{gray!15} 0.457 & \cellcolor{gray!15} 0.721 & \cellcolor{gray!15} 57.08 & \cellcolor{gray!15} 3.566 \\
    & \gls{model}-rbf & $\Theta_{preproc}$ & 111.61 & 0.831 & 0.075 & 37.53 & 4.739 \\
    & \cellcolor{gray!15} \gls{model}-rbf & \cellcolor{gray!15} $\Theta_{R_L}, \Theta_{pred}$ & \cellcolor{gray!15} 46.23 & \cellcolor{gray!15} 0.324 & \cellcolor{gray!15} 0.859 & \cellcolor{gray!15} 70.30 & \cellcolor{gray!15} 1.911 \\
    & \cellcolor{gray!15} \gls{model}-rbf & \cellcolor{gray!15} $\Theta_{V_L}, \Theta_{pred}$ & \cellcolor{gray!15} 58.13 & \cellcolor{gray!15} 0.409 & \cellcolor{gray!15} 0.776 & \cellcolor{gray!15} 60.81 & \cellcolor{gray!15} 3.271 \\
    & \cellcolor{gray!15} \gls{model}-rbf & \cellcolor{gray!15} $\Theta_{preproc}, \Theta_{pred}$ & \cellcolor{gray!15} 56.70 & \cellcolor{gray!15} 0.399 & \cellcolor{gray!15} 0.787 & \cellcolor{gray!15} 61.94 & \cellcolor{gray!15} 2.765 \\
    & \cellcolor{gray!15} \gls{model}-rbf & \cellcolor{gray!15} $\Theta_{V_L}, \Theta_{R_L}, \Theta_{pred}$ & \cellcolor{gray!15} 45.07 & \cellcolor{gray!15} 0.317 & \cellcolor{gray!15} 0.866 & \cellcolor{gray!15} 71.19 & \cellcolor{gray!15} 1.745 \\
    & \cellcolor{gray!15} \gls{model}-rbf & \cellcolor{gray!15} $\Theta_{preproc}, \Theta_{R_L}, \Theta_{pred}$ & \cellcolor{gray!15} 44.78 & \cellcolor{gray!15} 0.316 & \cellcolor{gray!15} 0.866 & \cellcolor{gray!15} 71.70 & \cellcolor{gray!15} 1.698 \\
    & \cellcolor{gray!15} \gls{model}-rbf & \cellcolor{gray!15} $\Theta_{preproc}, \Theta_{V_L}, \Theta_{pred}$ & \cellcolor{gray!15} 56.98 & \cellcolor{gray!15} 0.402 & \cellcolor{gray!15} 0.784 & \cellcolor{gray!15} 61.52 & \cellcolor{gray!15} 3.019 \\
    & \cellcolor{gray!15} \gls{model}-rbf & \cellcolor{gray!15} $\Theta_{preproc}, \Theta_{V_L}, \Theta_{R_L}, \Theta_{pred}$ & \cellcolor{gray!15} 44.96 & \cellcolor{gray!15} 0.318 & \cellcolor{gray!15} 0.865 & \cellcolor{gray!15} 71.59 & \cellcolor{gray!15} 1.901 \\
    & \cellcolor{gray!15} \gls{model}-rbf & \cellcolor{gray!15} All & \cellcolor{gray!15} 42.74 & \cellcolor{gray!15} 0.303 & \cellcolor{gray!15} 0.877 & \cellcolor{gray!15} 73.64 & \cellcolor{gray!15} \textbf{1.531} \\
    \bottomrule
  \end{tabularx}%

  \scriptsize
  \vspace{1ex}
  Notes:\\Arrows indicate the desired direction of improvement for each metric.\\Bold values denote the best overall performance for a specific metric within each combination.\\Highlighted rows indicate configurations where the proposed architecture outperforms the baseline across all evaluation metrics, for the specific Experiment / Dataset combination.
  \caption{Results and statistics from Experiment D (Dataset B, Distribution Shift). Even when transferring across disparate networks under the challenging conditions of Dataset B, the \gls{model} architectures consistently match and outperform the baseline across most fine-tuning configurations.}
  \label{tab:exp-c_dataset_b-results}
\end{table}

In the following discussion, the different subsets of adapted parameters are defined as follows:
\begin{itemize}
  \item $\Theta_{preproc}$: the parameters of the initial node feature construction layers, responsible for processing the input demand and generating the initial node embeddings. To maintain notational brevity, this term denotes both the \gls{mlp}-based preprocessor in the baseline model, and the proposed initialization layers in the \gls{model} models.
  \item $\Theta_{V_L}$: the parameters of the final layer of the \gls{vencoder}.
  \item $\Theta_{R_L}$: the parameters of the final local topological routing layer of the \gls{rencoder}.
  \item $\Theta_{pred}$: the parameters of the edge-level flow predictor, which maps the final latent node and edge representations to scalar flow predictions.
\end{itemize}

The transferability under the random uniform scaling scenarios of Dataset A reveals the necessity of domain adaptation: zero-shot transfer fails completely, with negative $R^2$ values and severe physical flow violations. However, both the baseline and the proposed \gls{model} models demonstrate strong recovery under highly restricted parameter-efficient domain adaptation. Unfreezing the final layer of the real-edge message passing and the edge-level predictor ($\Theta_{R_L}, \Theta_{pred}$) already restores near-native predictive functionality. The \gls{model} models generally outperform the baseline in this restricted adaptation regime, only marginally underperforming the baseline in specific configurations (e.g. when unfreezing $\Theta_{R_L}, \Theta_{pred}$ on \gls{an-a} for \gls{model}-lin).

The same trend can be observed under the distribution shift scenarios from Dataset B, where again the \gls{model} models generally outperform the baseline across most transfer configurations, with improvements in both predictive accuracy and physical consistency. However, specific configurations show that the proposed models outperform the baseline model in some metrics, while performing slightly worse in others (e.g. \gls{model}-lin outperforming the baseline on all metrics except the \gls{fcn} metric on \gls{ch-b} when unfreezing $\Theta_{R_L}, \Theta_{pred}$, with 2.449 vs 2.405).

\paragraph{Interpretation}
The results of this sensitivity analysis provide valuable insights into the internal mechanics of how the tested architectures encode the \gls{tap} and the relative importance of different components for transferability across domains.

Our first findings highlight that zero-shot transfer fails catastrophically for both models. This confirms that the learned representations of traffic flow dynamics are highly specific to the structure of the training domain, and that without adaptation, the tested models cannot generalize to new network topologies. However, the \gls{model} architectures consistently demonstrate stronger or equal capacity for domain adaptation compared to the baseline model across all tested frozen parameter configurations.

Further analysis reveals that fine-tuning only the prediction heads ($\Theta_{pred}$) provides a significant improvement over zero-shot, but remains insufficient to achieve acceptable predictive accuracy or physical consistency. This suggests that while the prediction heads map latent representations to scalar flows, the underlying feature extraction layers must also be calibrated to capture the unique characteristics of the target network.

On both datasets, we observe that the critical performance breakthrough exclusively occurs when the local topological routing layers ($\Theta_{R_L}$) are adapted alongside the predictor layers ($\Theta_{pred}$). Furthermore, unfreezing the initialization layer in addition to the local routing layer and the predictor ($\Theta_{preproc}, \Theta_{R_L}, \Theta_{pred}$) does not yield significant performance improvements for any of the tested models. This suggests that the bottleneck in transfer learning performance does not lie in the demand projection itself, but rather in the downstream message-passing layers, which must be adapted to capture the unique topological geometries of the target network.

Even when computational constraints prevent full encoder fine-tuning, which is shown to provide the best adaptation to the target domain, the models equipped with the proposed \gls{layer} layer demonstrate a significant performance benefit over the baseline model across the majority of the tested configurations. In particular, fine-tuning just the final local receptive field prior to the scalar flow prediction, in the form of the last \gls{rencoder} layer and the prediction heads ($\Theta_{R_L}, \Theta_{pred}$), is sufficient to surpass 70\% GEH compliance across \gls{an-b} and \gls{ch-b} for both \gls{model} variants, a meaningful improvement over the baseline model which only reaches 63\% compliance in the same conditions. This highlights the increased parameter-efficiency deriving from the proposed initialization framework, which allows the models to achieve significant performance improvements with minimal fine-tuning of a small subset of parameters.

\subsection{Computational Efficiency}
\begin{figure}[hbtp]
  \centering
  \includegraphics[width=0.8\textwidth]{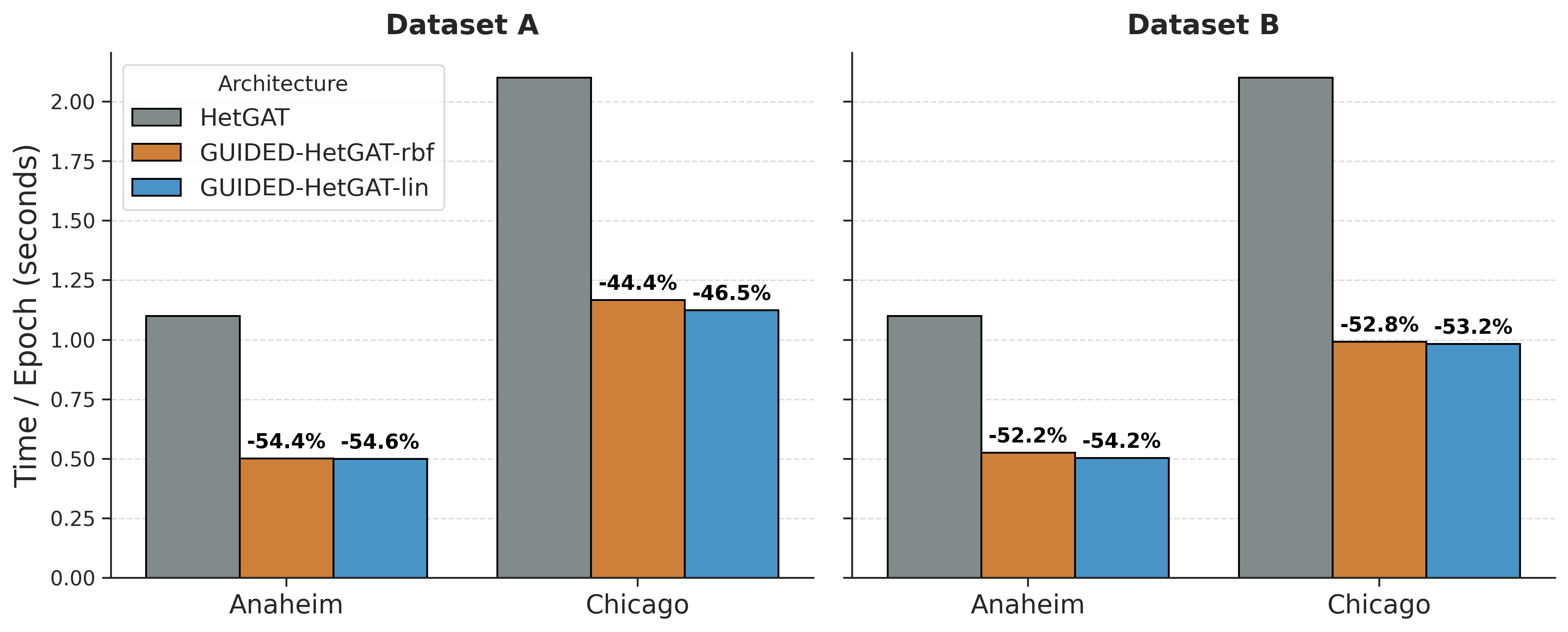}
  \captionof{figure}{Comparison of the computational efficiency of the proposed \gls{model} models against the baseline \gls{hetgat} on Dataset A (left) and Dataset B (right) in Experiment A.}
  \label{fig:computation-efficiency}
\end{figure}

The computational efficiency of the proposed \gls{model} models is evaluated in terms of training time per epoch, and compared against the baseline \gls{hetgat} on both datasets and networks for Experiment A, which is the most computationally demanding. \cref{fig:computation-efficiency} shows the average training time per epoch for each model and dataset.

Both \gls{model} variants achieve a significant reduction in training time per epoch compared to the baseline \gls{hetgat}, with reductions of around 50\% across all datasets and networks (ranging from 44.4\% to 54.6\%). This improvement can be attributed to the more efficient feature initialization process introduced by the \gls{layer} layer, which leverages highly optimized scatter operations for the node feature construction, as opposed to the more computationally intensive \gls{mlp}-based preprocessing of the baseline model. The consistent reduction in training time across all scenarios highlights the computational efficiency of the proposed initialization framework, which can provide significant speedups even in large-scale, complex traffic assignment tasks.

\section{Conclusions} \label{ch:conclusion}
This research set out to address a critical bottleneck in the application of deep learning to transportation engineering: the spatial transferability of data-driven surrogate models for the macroscopic \gls{tap}. By developing and rigorously evaluating a novel procedure for the initialization of node features from demand data, this work demonstrated how to successfully decouple travel demand representation from specific network topologies, enabling the development of truly inductive surrogate \gls{gnn} models for the \gls{tap}.

By shifting the initial demand representation from a fixed-dimension node feature matrix to a scalar edge attribute, the proposed initialization paradigm renders the downstream architecture completely agnostic to the size of the network. This contrasts with previous architectures, where enabling transductive architectures in cross-network scenarios requires homogenizing the input features of the networks via zero-padding to match the same dimensionality on all tested datasets, limiting their applicability to networks of a predefined maximum size. The proposed framework elegantly bypasses this step, allowing the message-passing layers to rely on uniform, rich, high-dimensional embeddings without artificial dimensionality constraints.

The initialization procedure was integrated into a state-of-the-art heterogeneous graph attention architecture (\gls{hetgat}), and systematically benchmarked against the original \gls{hetgat} across multiple urban topologies and demand conditions. The experimental framework was designed to evaluate not only raw predictive accuracy, but also sample efficiency, robustness to spatially uncorrelated demand patterns, adherence to physical flow conservation, and inter-network transferability. The results of this evaluation can be summarized into four core findings:
\begin{enumerate}
  \item \textbf{Structural advantage under out-of-distribution demand patterns and data scarcity}: The tested \gls{model} model variants consistently outperformed the baseline model across most tested scenarios, showing particular resilience when confronted with the challenging conditions of Dataset B (Distribution Shift).
  \item \textbf{Initialization strategy performance:} While both \gls{layer} initialization strategies performed comparably across most scenarios, achieving state-of-the-art predictive accuracy and physical consistency, empirical results indicate a slight but consistent performance edge for \gls{rbf}-based variant. The \gls{rbf} expansion likely provides a richer, linearly separable initial embedding space for the demand information, granting the downstream layers a superior representational foundation.
  \item \textbf{Domain-agnostic transferability}: While zero-shot transfer failed across all model-dataset combinations, the model integrated with the proposed initialization method exhibited a greater capacity for parameter-efficient domain adaptation, allowing for seamless transitions between networks of vastly different sizes without the need for artificial input modifications.
  \item \textbf{Higher computational efficiency}: The proposed initialization layer achieved significant reductions in training time per epoch compared to the baseline approach, with improvements of around 50\% across all tested datasets and networks, highlighting the computational efficiency of the proposed initialization procedure.
\end{enumerate}

While the proposed \gls{layer} layer successfully addresses the spatial transferability gap in data-driven traffic assignment, the findings of this study must be contextualized within the boundaries of several foundational assumptions and methodological limitations. These same limitations also highlight promising avenues for future research, spanning from algorithmic scalability to the integration of more complex traffic dynamics.

\paragraph{Scalability boundaries}
From an architectural and computational perspective, the study faces specific boundaries, primarily driven by the inherent scalability limits of heterogeneous graphs. While the inclusion of a dedicated virtual link for every active \gls{od} pair allows our framework to successfully decouple demand from the nodes, it also causes the spatial complexity of the graph to scale quadratically with the number of \glspl{taz} ($\mathcal{O}(|\mathcal{Z}|^2)$). This ia structural limitation, shared with similar heterogeneous \gls{gnn} frameworks. Although effective for medium-to-large topologies like Anaheim and Chicago Sketch, extending this paradigm to massive, detailed city-scale networks would introduce a significant memory bottleneck, requiring advanced graph sampling or edge-pruning techniques to remain computationally efficient.

Future research should explore heuristic or learned virtual edge pruning, such as discarding \gls{od} connections with negligible demand prior to message passing, to drastically reduce the memory footprint of the initialization module without sacrificing the physical accuracy of the macroscopic flow predictions. Additionally, exploring hierarchical \gls{gnn} representations, where local neighborhood demands are aggregated into regional super-nodes before virtual edge projection, could provide a computationally tractable pathway to enable further architecture scaling.

\paragraph{Extension to multi-modal, spatio-temporal \acrlongpl{gnn}}
The experimental framework fundamentally relies on the deterministic \gls{tap} solved via the \gls{bfw} algorithm to generate ground-truth data. Consequently, the methodology inherently assumes a static network state where demand matrices and link capacities remain constant over the analysis period. By targeting an aggregated, time-independent equilibrium, the models abstract away complex temporal traffic dynamics, such as the formation and dissipation of transient shockwaves, queue spillbacks, and time-varying departure choices. Furthermore, this approach relies on Wardrop's First Principle, implicitly assuming that all network users belong to a homogeneous vehicle fleet, possess perfect information regarding network conditions, and behave perfectly rationally to minimize their individual travel times. The effects of multi-modal interactions, public transit schedules, and the distinct routing behaviors of heavy goods vehicles are not incorporated into the underlying link capacity or travel time functions.

While \gls{gnn} models have already been successfully extended to multi-modal and spatio-temporal settings in previous work \citep[e.g.][]{liu2025multi,ijcai2019p264,jiang2025deep}, the proposed network-agnostic initialization procedure has not yet been adapted to these more complex settings. Future research should explore how to extend the proposed framework to multi-modal and spatio-temporal \glspl{gnn}, incorporating time-varying demand and mode-specific features, allowing the model to capture the complex interactions between different transportation modes and their impact on overall network performance, while simultaneously accounting for non-static demand fluctuations and transient traffic dynamics.

\paragraph{Adaptation for empirical traffic observations}
Relying on synthetic, algorithmically generated ground-truth flow data allowed for a controlled evaluation of the proposed architecture, ensuring that the models were trained and tested on mathematically consistent equilibrium solutions. However, this approach abstracts away real-world complexities such as measurement noise, incomplete network observability, and non-equilibrium traffic conditions. Future research should validate the proposed architecture against empirical traffic data from GPS trajectories, induction loops, and camera-based sensors to assess its robustness and predictive accuracy in real-world scenarios.

\paragraph{Leveraging the virtual embeddings}
While the proposed architecture successfully decouples demand representation from the network topology, the current implementation only utilizes the virtual edge embeddings during the initialization phase, and does not leverage them in the subsequent message-passing layers, as detailed in \cref{sec:model-integration}.

Future research should explore how to effectively integrate these demand-informed virtual edge features into the message-passing process, allowing the model to dynamically incorporate demand information at each layer of the network, potentially enhancing its ability to capture complex demand-flow interactions and further improving predictive accuracy and physical consistency. In the specific case of the application to the \gls{hetgat} architecture, the virtual edge features could be used to modulate the attention weights during message passing in the \gls{vencoder}, allowing the model to dynamically adjust the influence of different nodes and edges based on the underlying demand patterns.

\paragraph{Extension to other domains}
Finally, while the proposed architecture is demonstrated in the context of vehicular traffic assignment, its fundamental design principles are not inherently tied to this specific application. By decoupling spatial topologies from demand matrices, the framework offers a versatile blueprint for surrogate modeling of any spatial interaction problem driven by \gls{od} matrices and other similar features, such as freight logistics and multimodal network optimization.

Future research should explore the application of this network-agnostic architecture to other domains, assessing its adaptability and performance in different contexts and with different types of spatial interaction data.

\section*{Acknowledgements}
This research has been supported by European Union's Horizon Europe programme under grant agreement No. 101138449 [project MI-TRAP (MItigating Transport-Related Air Pollution)], and by the TUM School of Social Sciences and Technology and the TUM Think Tank through the Friedrich Schiedel Fellowship under grant agreement No. FSF309 [project FAIR-NET (Fairness in Metropolitan Transport Networks)].

\bibliography{references.bib}

\end{document}